\documentclass[pdflatex,sn-nature]{sn-jnl}% Style for submissions to Nature Portfolio journals

\usepackage[utf8]{inputenc}%
\usepackage{graphicx}%
\graphicspath{{./}{figures/}}%
\usepackage{multirow}%
\usepackage{amsmath,amssymb,amsfonts}%
\usepackage{amsthm}%
\usepackage[title]{appendix}%
\usepackage{xcolor}%
\usepackage{textcomp}%
\usepackage{manyfoot}%
\usepackage{booktabs}%
\usepackage{array}%
\usepackage{tabularx}%
\usepackage{ragged2e}%
\usepackage{algorithm}%
\usepackage{algorithmicx}%
\usepackage{algpseudocode}%
\usepackage{listings}%
\hypersetup{hypertexnames=false}%

\theoremstyle{thmstyleone}%

\theoremstyle{thmstyletwo}%

\theoremstyle{thmstylethree}%

\begin{document}

\title[Evolutionary Intelligence for Scientific Discovery]{Evolutionary Intelligence for Scientific Discovery: From Evolutionary Computation to Cumulative Discovery Systems}

\author[1]{\fnm{Chao} \sur{Wang}}\email{xiaofengxd@126.com}
\author[1]{\fnm{Lingling} \sur{Li}}\email{llli@xidian.edu.cn}
\author[1]{\fnm{Fang} \sur{Liu}}\email{f63liu@163.com}
\author*[1]{\fnm{Licheng} \sur{Jiao}}\email{lchjiao@mail.xidian.edu.cn}
\affil*[1]{\orgdiv{School of Artificial Intelligence}, \orgname{Xidian University}, \orgaddress{\street{No. 2 South Taibai Road}, \city{Xi'an}, \postcode{710071}, \state{Shaanxi}, \country{China}}}

\abstract{
Artificial intelligence (AI) is shifting scientific discovery from task-specific workflows towards autonomous systems that organize exploration with experimental and human feedback in open-ended candidate spaces. Evolutionary computation (EC) provides a computational basis for feedback-driven discovery because population-based search can maintain diverse scientific candidates while steering exploration through accumulated evidence. However, EC predominantly focuses on candidate refinement for predefined problems, whereas cumulative discovery requires experience retention. To bridge this gap, this review introduces evolutionary intelligence (EI) for scientific discovery. EI characterizes scientific AI systems that sustain exploration by linking candidate refinement with experience retention across evolutionary cycles. We introduce a five-dimensional analytical framework that asks what evolves, how candidates change, why candidates are selected, where feedback originates, and when evolution occurs. This framework clarifies how EI transforms isolated search trajectories into cumulative scientific insight. We further demonstrate this paradigm across diverse discovery modes, from evolving concrete scientific entities to orchestrating automated research workflows. Finally, we identify critical bottlenecks regarding evaluation, process traceability, and shared infrastructure, providing a concrete roadmap for advancing the transition from EC to EI in scientific discovery.
}


\maketitle

\section{Introduction}\label{sec:introduction}

% 人工智能（AI）正日益成为科学发现的通用工具，其应用范围从特定任务的自动化，扩展至组织科学探索的自主系统 \cite{wang2023scientific,hey2009fourth}。通过利用实验反馈与人类交互，AI 系统能够以多轮循环的方式探索科学问题。这种范式转变在多个学科中已得到印证：AI 不再局限于被动的数据分析，而是主动在分子与材料设计 \cite{butler2018machine,merchant2023scaling}、蛋白质工程 \cite{jumper2021highly} 以及自动化实验室 \cite{king2009automation} 中探测候选空间。
Artificial intelligence (AI) increasingly serves as a general-purpose tool for scientific discovery, transitioning from task-specific automation to systems that organize scientific exploration \cite{wang2023scientific,hey2009fourth}. By leveraging experimental feedback and human interaction, AI systems can explore scientific problems through discovery cycles. This shift is evident across disciplines, where AI actively investigates candidate spaces in molecular and materials design \cite{butler2018machine,merchant2023scaling}, protein engineering \cite{jumper2021highly}, and automated laboratories \cite{king2009automation,boiko2023autonomous}, moving beyond passive data analysis.

% 反馈驱动的科学发现致力于探索边界随证据积累而逐渐显现的开放式候选空间 \cite{wang2023scientific}。由于实验反馈成本高昂且受物理条件约束，每一个被评估的候选对象（包括失败的尝试）都构成了宝贵的科学证据 \cite{macleod2020self,granda2018controlling}。因此，评估历史能够揭示候选空间内的隐藏结构，并支持多样化的探索行为 \cite{messeri2024artificial,kudela2022critical}。
Feedback-driven discovery explores open-ended candidate spaces where boundaries are often undefined and emerge as evidence accumulates \cite{lehman2024evolution,wang2023scientific}. Because experimental feedback is costly and physically constrained, every evaluated candidate, including failures, constitutes valuable scientific evidence \cite{macleod2020self,granda2018controlling}. Consequently, the evaluation history can reveal hidden structures within the candidate space and support diverse exploratory behaviours \cite{messeri2024artificial,lehman2024evolution}.

% 演化计算（EC）为驾驭这些复杂的发现周期提供了基础性的计算框架。与依赖单点搜索不同，EC 维护着一个科学候选种群，这些候选对象会根据反馈通过演化算子进行迭代更新。这种基于种群的探索方法特别适用于开放式候选空间，这类空间通常具有梯度缺失、反馈存在噪声以及边界动态演变的特征 \cite{salimans2017evolution,jin2021data,kudela2022critical,sarkar2025hyperscale}。通过将探索引导至有累积证据支持的区域，EC 能够在多个发现周期中持续生成多样化的科学候选对象。这种能力涵盖了广泛的算法传统，从经典的遗传算法和进化策略 \cite{fogel1966artificial,rechenberg1973evolutionsstrategie,holland1975adaptation,goldberg1989genetic,koza1992genetic}，到现代的自然进化策略 \cite{hansen2001completely,wierstra2014natural}，再到质量多样性搜索方法 \cite{lehman2011abandoning,mouret2015illuminating,pugh2016quality,stanley2017openendedness}。
Evolutionary computation (EC) provides a foundational computational framework to support these discovery cycles \cite{eiben2015evolution,fogel1966artificial}. Rather than relying on a single-point search, EC maintains populations of scientific candidates that are iteratively updated by evolutionary operators according to feedback. This population-based approach is particularly well suited to open-ended candidate spaces characterized by weak or absent gradients, noisy feedback, and dynamically evolving boundaries \cite{lehman2024evolution,salimans2017evolution,jin2021data,kudela2022critical}. By steering exploration toward regions supported by accumulated evidence, EC sustains the creation of diverse scientific candidates across discovery cycles. This approach encompasses various common algorithms, such as classical genetic algorithm, evolution strategies \cite{holland1975adaptation,goldberg1989genetic,koza1992genetic,rechenberg1973evolutionsstrategie}, natural evolution strategies \cite{hansen2001completely,wierstra2014natural}, and quality-diversity search methods \cite{lehman2011abandoning,mouret2015illuminating,pugh2016quality,stanley2017openendedness}.

% 传统上，进化计算（EC）主要通过优化针对明确定义的科学问题的候选方案来推进人工智能在科学发现领域的应用。大量研究致力于设计专门的进化算子来引导这种候选方案的优化过程[Holland 1975 Adaptation, Goldberg 1989 Genetic, Koza 1992 Genetic, Hansen 2001 Completely]。然而，累积性发现不仅仅依赖于找到最佳候选方案。整个搜索轨迹，包括失败的尝试和复杂的反馈，都能提供指导后续搜索的经验[Cully 2015 Robots, Mouret 2015 Illuminating, Gupta 2021 Evolutionary, Stanley 2017 Openendedness]。因此，进化搜索不仅用于解决单个问题，还用于在迭代发现周期中积累经验[Wang 2026 LLM EA, Self-Evolving Agents 2026]。
Conventionally, EC has advanced AI for scientific discovery primarily by refining candidates for well-defined scientific problems. Extensive research has focused on designing specialized evolutionary operators to steer this candidate refinement \cite{holland1975adaptation,goldberg1989genetic,koza1992genetic,hansen2001completely}. However, cumulative discovery relies on more than just finding the best candidate. The entire search trajectory, including failed trials and complex feedback, provides experience that guides later searches \cite{cully2015robots,mouret2015illuminating,gupta2021evolutionary,stanley2017openendedness}. Therefore, evolutionary search serves not just to solve single problems, but to accumulate experience across iterative discovery cycles \cite{wang2026llmEA,selfevolvingagents2026}.

% 为了将这一范式转变形式化，本综述引入了面向科学发现的演化智能（EI）。EI 刻画了那些通过将候选精炼与跨演化周期的经验保留内在结合，从而维持探索活动的科学 AI 系统。在该框架内，科学候选种群由演化算子更新，并通过实验或人类反馈进行评估。关键在于，这些已评估的候选对象及其对应的反馈会以结构化形式被保留，从而指导后续的探索。与主要针对孤立问题最大化样本效率的单任务搜索方法不同 \cite{shahriari2016taking,frazier2018tutorial,settles2009active,shields2021bayesian}，EI 将整个搜索轨迹（包括失败的尝试和候选谱系）视为首要的科学证据，以便进行结构化和跨周期重用。通过将候选精炼与经验保留相连接，EI 为累积式发现提供了坚实的框架 \cite{holland1975adaptation,goldberg1989genetic,hansen2001completely,salimans2017evolution,deb2002fast,mouret2015illuminating,cully2015robots,gupta2021evolutionary,wang2026llmEA,selfevolvingagents2026}。
To capture this transition, this review introduces evolutionary intelligence (EI) for scientific discovery. EI characterizes scientific AI systems that sustain exploration by linking candidate refinement with experience retention across evolutionary cycles. Within this framework, populations of scientific candidates are updated by evolutionary operators and evaluated via experimental or human feedback. Specifically, these evaluated candidates and their corresponding feedback are retained in structured forms to guide subsequent exploration. Unlike single-task search methods that primarily maximize sample efficiency for isolated problems \cite{jin2021data,frazier2018tutorial,settles2009active}, EI treats the entire search trajectory, including failed trials and candidate lineages, as scientific evidence to be structured and reused. By bridging candidate refinement with experience retention, EI provides a structured framework for cumulative discovery \cite{cully2015robots,wang2026llmEA,selfevolvingagents2026}.

% 近期的 AI 系统正日益将演化循环深度嵌入科学发现之中。算法发现平台（如 FunSearch 和 AlphaEvolve）通过自动化评估机制来精炼候选程序 \cite{romera2024mathematical,novikov2025alphaevolve}。基于基础模型的辅助演化方法进一步将这一范围从最终的科学目标扩展至基础的研究组件。这些系统主动演化提示词、代码和假设，以及可复用的技能 \cite{fernando2023promptbreeder,lehman2024evolution,gepa2026,oh2025discovering}。与此同时，自主实验室与科学智能体将探索过程与实验及人类反馈紧密耦合，使后续候选方案能够依据累积证据进行动态调整 \cite{burger2020mobile,agentgym2024,dgm2026,ghareeb2026multi,gottweis2026accelerating}。此类完全自主的研究生态系统的涌现，正日益被概念化为“代理科学”（agentic science） \cite{agentic_science2025,scientific_intelligence2025,ai4research2025,arxiv2410_05080,arxiv2510_14150}。随着演化搜索开始重塑这些研究组件，评估指标必须相应地超越单纯的候选质量，明确将科学可靠性与人类监督纳入核心考量 \cite{akiba2025evolutionary,skillrl2026,agent02025}。
Recent AI systems increasingly embed evolutionary cycles into scientific discovery. Algorithm-discovery platforms, such as FunSearch and AlphaEvolve, refine candidate programs through automated evaluation mechanisms \cite{romera2024mathematical,novikov2025alphaevolve}. Foundation-model-assisted evolutionary methods further expand this scope beyond final scientific targets to fundamental research components. These systems actively evolve prompts, code, and hypotheses, alongside reusable skills \cite{fernando2023promptbreeder,lehman2024evolution,gepa2026,oh2025discovering,zhang2026evom,skillrl2026}. Concurrently, autonomous laboratories and scientific agents integrate exploration with experimental and human feedback, enabling subsequent candidates to adapt dynamically to accumulated evidence \cite{burger2020mobile,macleod2020self,boiko2023autonomous,agentgym2024,dgm2026,ghareeb2026multi,gottweis2026accelerating}. The emergence of such autonomous research systems is increasingly recognized as agentic science \cite{agentic_science2025,scientific_intelligence2025,ai4research2025,arxiv2410_05080,arxiv2510_14150}. As evolutionary search begins to reshape these research components, evaluation metrics need to expand beyond mere candidate quality to explicitly incorporate scientific reliability and human oversight \cite{akiba2025evolutionary,skillrl2026,agent02025,lu2026automation,messeri2024artificial}.

% 本综述探讨了演化计算（EC）向面向科学发现的演化智能（EI）的转变。我们引入了一个五维分析框架，以系统地解构多样化的科学 AI 系统，涵盖从基础模型辅助的程序搜索到闭环自主实验室。该框架阐明了 EI 循环如何将孤立的搜索经验转化为结构化的科学洞见。我们进一步展示了这一机制如何在多样化的发现模式中运作，从演化具体实体到编排自动化工作流。最后，我们概述了评估、可追溯性和基础设施方面的关键挑战，并为推动从 EC 向 EI 的转型提供了具体的路线图。最终，这一视角将 EC 从任务级的搜索工具重塑为累积式科学发现的组织原则。
This review examines the transition from EC to EI in scientific discovery. We first introduce a five-dimensional analytical framework designed to analyze diverse scientific AI systems, from foundation-model-assisted program search to closed-loop autonomous laboratories. This framework clarifies how the EI cycle transforms isolated search experiences into structured scientific insights. We further demonstrate how this paradigm operates across diverse discovery modes, from evolving concrete scientific entities to orchestrating automated research workflows. Finally, we outline key challenges regarding evaluation, process traceability, and shared infrastructure, providing a concrete roadmap to advance this transition. Ultimately, this perspective reframes EC from a task-level search tool into an organizing principle for cumulative scientific discovery.

\section{From evolutionary computation to evolutionary intelligence}\label{sec:ec-to-ei}

EC provides a foundational framework for population-based search in scientific discovery. Rather than relying on single-point updates, EC maintains populations of candidates through evolutionary operators that organize reproduction, evaluate outcomes against feedback, and drive selection. Adaptive strategies and archival mechanisms retain historical experience to influence subsequent search trajectories \cite{fogel1966artificial,rechenberg1973evolutionsstrategie,holland1975adaptation,goldberg1989genetic,hansen2001completely}. This organizational logic is well suited to open-ended scientific candidate spaces. In such environments, search often proceeds before boundaries, objectives, or feedback processes are fully defined, aligning with broader biological perspectives on adaptation and the evolution of physical artifacts \cite{cussatblanc2020biological,eiben2015evolution}. By preserving diverse candidate lineages and leveraging quality-diversity principles, EC sustains robust exploration even under weak gradients or noisy feedback \cite{mouret2015illuminating,pugh2016quality,stanley2017openendedness,salimans2017evolution,jin2021data,sarkar2025hyperscale}.

While EC provides the search mechanism, many traditional applications treat scientific discovery as black-box optimization over fixed representations. Unlike these approaches, EI broadens the scope of the evolving object. Candidate populations in EC need not be stored exclusively as explicit sets. While genetic algorithms \cite{goldberg1989genetic} and genetic programming \cite{koza1992genetic} traditionally maintain explicit candidate representations, natural evolution strategies \cite{hansen2001completely,wierstra2014natural} and estimation-of-distribution algorithms \cite{larranaga2002estimation,rubinstein1999cross,mockus1978application} update probabilistic sampling distributions derived from evaluated candidates. In scientific discovery, evolution can act on scientific targets \cite{jumper2021highly,merchant2023scaling}, foundational research components \cite{stanley2019designing,jaderberg2017population,romera2024mathematical,fernando2023promptbreeder,gepa2026}, up to the discovery processes themselves \cite{burger2020mobile,macleod2020self,agentgym2024,voyager2023}. For instance, rather than merely refining a molecular structure, an EI system might concurrently evolve the machine learning model predicting its properties, alongside the experimental batching policy used to test it \cite{chu2026programmable,ghareeb2026multi,gottweis2026accelerating}. This shift from fixed representations to dynamic, multi-level targets allows scientific AI systems to organize exploration flexibly, whether through explicit candidate pools and learned generative models, alongside probabilistic proposal distributions. However, as the evolving objects become more complex and scientifically grounded, the evaluation mechanism needs to evolve beyond simple scalar fitness assignment.

Scientific feedback expands the role of evaluation beyond simple ranking to incorporate diverse physical and computational constraints. In EC practices, evaluation is typically reduced to a scalar objective value used to rank candidates \cite{holland1975adaptation,goldberg1989genetic,back1996evolutionary}. In scientific discovery, however, feedback encompasses various types of signals. These range from data-driven computational proxies, such as predictive models and uncertainty estimates \cite{jin2021data,butler2018machine,jumper2021highly,merchant2023scaling}, to physical constraints derived from laboratory experiments \cite{burger2020mobile,macleod2020self,boiko2023autonomous,granda2018controlling}, alongside human-driven reproducibility checks and expert judgement \cite{wang2023scientific,messeri2024artificial,hey2009fourth}. Consequently, these diverse signals shape not only which candidates are selected, but also how the underlying models, constraints, and experimental priorities are adjusted for the subsequent discovery cycle. These signals generate rich scientific evidence across every cycle, which raises the key question of how such evidence is preserved and utilized over time.

\begin{figure}[t]
\centering
\includegraphics[width=0.98\textwidth]{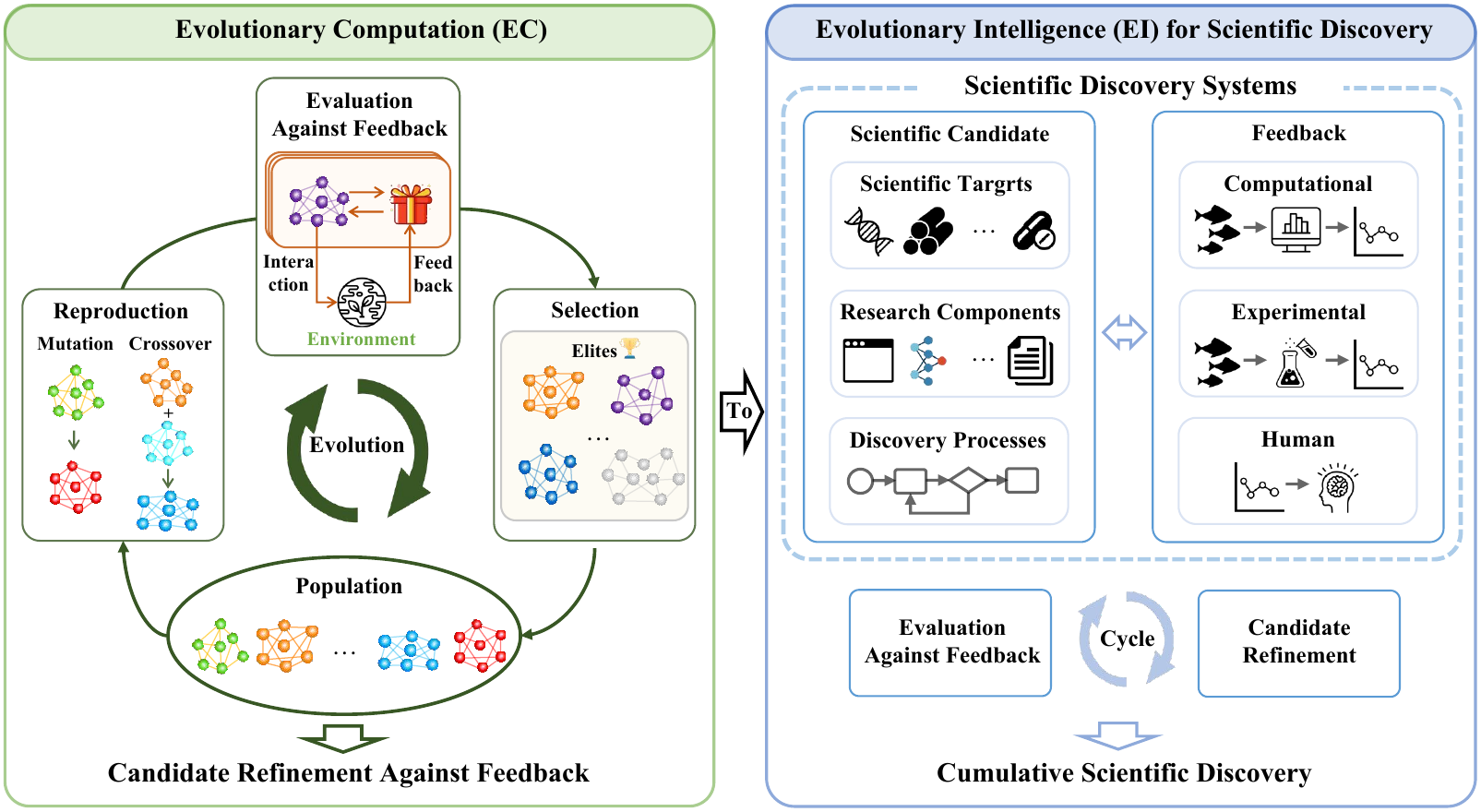}
\caption{From evolutionary computation (EC) to evolutionary intelligence (EI) for scientific discovery. EC organizes population-based search through evolutionary operators. EI embeds this organization in scientific discovery systems where evolution can act on scientific targets, research components, or discovery processes. The transition from EC to EI shifts the emphasis from candidate refinement against feedback to cumulative scientific discovery.}
\label{fig:ec-to-ei}
\end{figure}

\begin{table}[t]
\caption{From evolutionary computation to evolutionary intelligence for scientific discovery.}\label{tab:ec-ei}
\small
\begin{tabularx}{\textwidth}{@{}p{0.20\textwidth}XX@{}}
\toprule
\textbf{Dimension} & \textbf{Evolutionary computation} & \textbf{Evolutionary intelligence for scientific discovery} \\
\midrule
Primary role & Population-based search & Cumulative scientific discovery \\
Population & Candidate sets or sampling distributions & Scientific targets, research components, or discovery processes \\
Reproduction & Mutation, crossover, distributional sampling, and program variation & Evolutionary variation, learned generation, foundation-model generation, and human-guided modification \\
Feedback-based evaluation & Fitness functions, objective values, or simulations & Computational, experimental, and human feedback \\
Selection and update & Candidate survival and distribution update & Updated candidates, proposal distributions, models, constraints, and experimental priorities \\
Archive and record & Elites, non-dominated solutions, diverse behaviours, and search states & Successful candidates, failures, lineages, model states, experimental records, and annotations \\
Experience retention and transfer & Adaptation within a search loop and transfer across related tasks & Experience reuse across evolutionary cycles and transfer across tasks, models, skills, or experiments \\
\botrule
\end{tabularx}
\end{table}

Ultimately, the integration of expanded evolving objects and diverse scientific feedback requires a shift in the temporal scale of experience retention, marking a key transition from isolated search to cumulative scientific discovery. Traditional adaptive or transfer methods typically confine experience utilization within a single search loop \cite{brest2006self} or between closely related tasks \cite{gupta2016multifactorial,wang2022solving}. By contrast, EI systematically archives the search trajectory, treating failed trials, candidate lineages, and experimental logs as valuable scientific evidence. This structured record reshapes candidate generation and selection pressures in subsequent cycles, enabling knowledge transfer across distinct tasks and scientific domains \cite{pan2010survey,gupta2021evolutionary,finn2017model,yosinski2014transferable}, as well as simulation-to-experiment boundaries \cite{wang2022solving}. Recent foundation-model-assisted agents \cite{romera2024mathematical,novikov2025alphaevolve,fernando2023promptbreeder,gepa2026,lehman2024evolution} and autonomous laboratories \cite{burger2020mobile,macleod2020self,agentgym2024} exemplify this paradigm. In these systems, validation, traceability, and human oversight operate as core elements of the EI framework rather than external checks \cite{lu2026automation,aygun2026ai,messeri2024artificial}. Fig.~\ref{fig:ec-to-ei} and Table~\ref{tab:ec-ei} summarize this transition from EC to EI for scientific discovery. Fig.~\ref{fig:ec-to-ei} illustrates how population-based search becomes embedded within feedback-driven discovery systems, while Table~\ref{tab:ec-ei} compares their core operations.

\section{A five-dimensional framework for evolutionary intelligence}\label{sec:architecture}

\begin{figure}[t]
\centering
\includegraphics[width=0.98\textwidth]{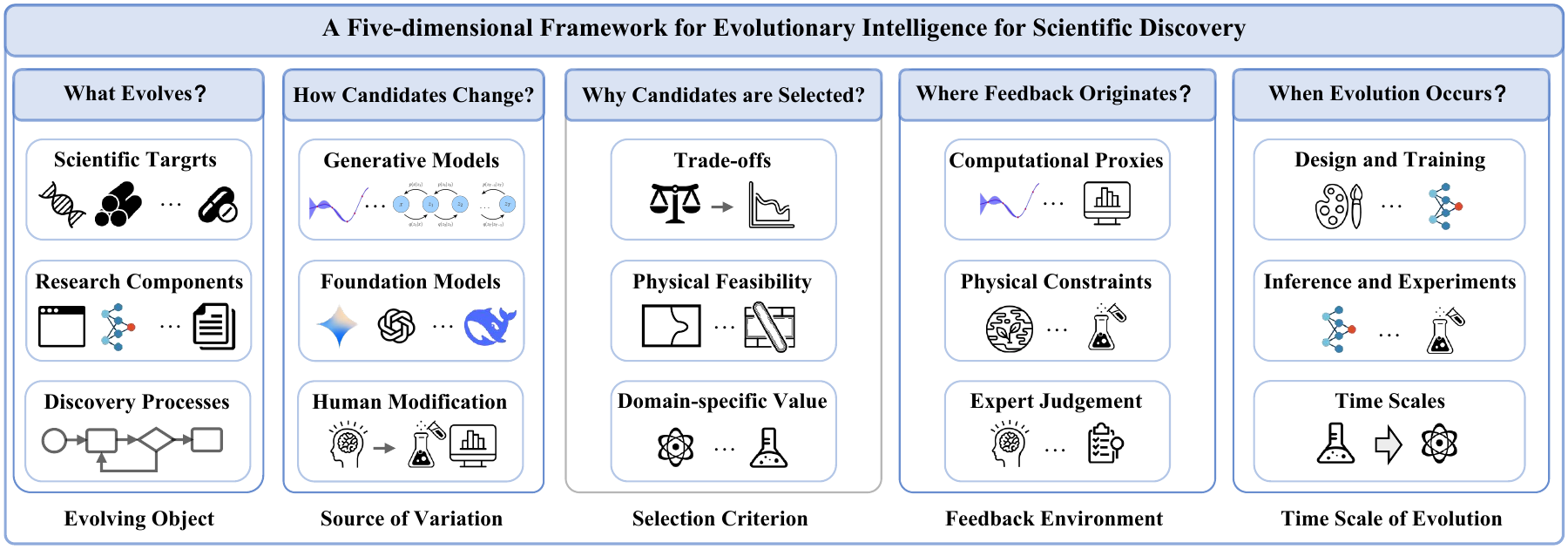}
\caption{A five-dimensional framework for evolutionary intelligence. This framework analyzes scientific AI systems by linking five analytical questions to their corresponding system components: the evolving object (what evolves), the source of variation (how candidates change), the selection criterion (why candidates are selected), the feedback environment (where feedback originates), and the time scale of evolution (when evolution occurs).}
\label{fig:architecture}
\end{figure}
% 图 2 演化智能的五维分析框架。该框架通过将五个分析问题映射到相应的系统组件来解构科学 AI 系统：演化对象（什么在演化）、变异来源（候选如何变化）、选择标准（为何选择候选）、反馈环境（反馈源自何处）以及演化的时间尺度（演化何时发生）。这些维度共同构成了一个控制累积式发现的闭环架构。

% 面向科学发现的演化智能（EI）可以通过五个相互关联的问题进行系统性解构，这五个问题共同构成了一个闭环反馈架构：什么在演化、候选如何变化、为何选择候选、反馈源自何处，以及演化何时发生。这些问题是映射到发现系统的五个紧密耦合的组件上——演化对象、变异来源、选择标准、反馈环境以及演化的时间尺度 \cite{holland1975adaptation}。关键在于，这些组件协同运作：反馈环境决定了选择标准，进而驱动变异机制，而时间尺度则掌控着累积经验如何跨周期被保留与迁移 \cite{gupta2021evolutionary}。这一统一框架（如图 2 所示）提供了一个诊断透镜，用于分析孤立的搜索轨迹如何转化为累积的科学洞见。
EI for scientific discovery can be analyzed through five interconnected questions: what evolves, how candidates change, why candidates are selected, where feedback originates, and when evolution occurs. These questions correspond to five components of a discovery system: the evolving object, the source of variation, the selection criterion, the feedback environment, and the time scale of evolution \cite{holland1975adaptation}. These components interact closely. The feedback environment influences the selection criteria, which in turn shape the mechanisms of variation, while the time scale determines how accumulated experience is retained and transferred across cycles \cite{gupta2021evolutionary}. This framework, illustrated in Fig.~\ref{fig:architecture}, provides a structured approach to analyze how isolated search trajectories are transformed into cumulative scientific insight.

% 演化是什么？许多进化计算（EC）应用通常针对固定表示，范围从数值向量到静态程序和预定义策略[1975年 Holland 适应, 1989年 Goldberg 遗传, 1992年 Koza 遗传, 2003年 Schmidhuber Godel, 2002年 Stanley 演化]。对于科学发现，进化计算（EI）将这一范围扩展到多个层面。演化的对象可以包含具体的科学目标、研究组成部分或发现过程。具体而言，目标范围从分子结构和蛋白质序列到复杂的实验方案[1971年 Jumper 高度, Merchant 规模化, Watson 广泛, Dauparas 稳健]。研究组成部分包括计算模型和数据表示[Gilmer 2017neural, Schutt 2018schnet, Batatia 2022mace]，以及提示模板和可检验的假设[Fernando 2023promptbreeder, Gepa 2026, Yao 2023tree]。此外，发现过程本身也会不断演进，包括实验规划方案、工具使用流程以及人机协作工作流程[Burger 2020mobile, Boiko 2023autonomous, Agentgym 2024, Voyager 2023]。这种更广阔的视角很有价值，因为这些对象在实践中经常相互重叠。一个科学人工智能系统通常会在一个周期内同时调整多个组件。例如，一个自主材料平台可以同时优化候选化合物、更新底层替代模型并调整实验批次策略\cite{jin2021data,chu2026programmable,ghareeb2026multi,gottweis2026accelerating}。类似地，可重用技能的演进，例如自动化数据管理和文献检索，作为一个动态的研究组成部分，支持更广泛的发现工作流程\cite{voyager2023,skillrl2026}。
\textbf{What evolves.} Many EC applications typically target fixed representations, ranging from numerical vectors to static programs and predefined policies \cite{holland1975adaptation,goldberg1989genetic,koza1992genetic,schmidhuber2003godel,stanley2002evolving}. For scientific discovery, EI broadens this scope across multiple levels. The evolving object may encompass concrete scientific targets, research components, or discovery processes. Specifically, targets range from molecular structures and protein sequences to complex experimental recipes \cite{jumper2021highly,merchant2023scaling,watson2023broadly,dauparas2022robust}. Research components include computational models and data representations \cite{gilmer2017neural,schutt2018schnet,batatia2022mace}, as well as prompt templates and testable hypotheses \cite{fernando2023promptbreeder,gepa2026,yao2023tree}. Furthermore, discovery processes themselves can evolve, including experimental planning protocols, tool-use routines, and human-AI collaborative workflows \cite{burger2020mobile,boiko2023autonomous,agentgym2024,voyager2023}. This broader view is valuable because these objects frequently overlap in practice. A scientific AI system often adapts multiple components concurrently within a single cycle. For instance, an autonomous materials platform might simultaneously refine candidate compounds, update underlying surrogate models, and adjust experimental batching policies \cite{jin2021data,chu2026programmable,ghareeb2026multi,gottweis2026accelerating}. Similarly, the evolution of reusable skills, such as automating data curation and literature retrieval, functions as a dynamic research component that supports the broader discovery workflow \cite{voyager2023,skillrl2026}.

% 候选如何变化。 在经典的 EC 中，候选变异主要由随机的繁殖算子驱动，例如突变与交叉，以及分布采样和程序变换 \cite{goldberg1989genetic,koza1992genetic,hansen2001completely}。在面向科学发现的 EI 中，这种变异机制通过学习到的生成模型、基础模型和人类引导的修改得到了显著丰富。例如，大型模型可以提出或修订复杂的程序与可测试的假设，同时完善提示词模板与实验计划 \cite{romera2024mathematical,fernando2023promptbreeder,gepa2026,lehman2024evolution,schick2023toolformer,novikov2025alphaevolve}。关键在于，特定领域的科学约束与先验知识起到了引导作用，将这种广泛的变异限制在物理或逻辑上合理的候选空间区域内。
\textbf{How candidates change.} EC algorithms primarily drive candidate variation through operators such as mutation and crossover, as well as distributional sampling and program transformation \cite{goldberg1989genetic,koza1992genetic,hansen2001completely,larranaga2002estimation}. In EI for scientific discovery, candidate variation is also driven by learned generative models, foundation models, and human-guided modification. For instance, large foundation models can propose or revise complex programs and testable hypotheses, and refine prompt templates and experimental plans \cite{romera2024mathematical,fernando2023promptbreeder,gepa2026,lehman2024evolution,schick2023toolformer,novikov2025alphaevolve,madaan2023selfrefine}. Domain-specific scientific constraints and prior knowledge can help guide this variation, restricting it to physically or logically plausible regions of the candidate space \cite{batatia2022mace,akiba2025evolutionary}.

% 候选者被选中的原因。许多进化计算（EC）实现依赖于标量适应度值或预定义的目标函数来对候选者进行排序和传播[参考文献1975：适应，Goldberg 1989：遗传，Back 1996：进化]。然而，在科学发现中，选择标准扩展到涵盖多维权衡、物理可行性和特定领域的科学价值。选择压力并非优先考虑单一指标，而是平衡相互竞争的目标，例如预测准确性、计算成本和实验安全性[参考文献1002：快速，Zhang 2007：Moead]。此外，保持新颖性和多样性的机制会惩罚过早收敛，从而保持备选假设和候选空间中未探索区域的可行性[参考文献11：放弃，Mouret 2015：启发，Pugh 2016：质量，Cully 2015：机器人，Stanley 2017：开放性]。这些标准会随着反馈的积累而调整，同时，人类的判断和专家的监督可以引导选择方向，使其指向那些缺乏直接计算验证但具有科学前景的区域[参考文献：settles2009active,gelbart2014bayesian,kandasamy2017multi]。因此，EI中的选择功能不仅在于筛选候选对象，还在于引导搜索方向朝着不断变化的科学优先事项发展。
\textbf{Why candidates are selected.} Many EC implementations rely on scalar fitness values or predefined objective functions to rank and propagate candidates \cite{holland1975adaptation,goldberg1989genetic,back1996evolutionary}. In scientific discovery, however, selection criteria expand to encompass multi-dimensional trade-offs, physical feasibility, and domain-specific scientific value. Rather than prioritizing a single metric, selection pressures balance competing objectives, such as predictive accuracy, computational cost, and experimental safety \cite{deb2002fast,zhang2007moead}. Furthermore, mechanisms that preserve novelty and diversity penalize premature convergence, keeping alternative hypotheses and unexplored regions of the candidate space viable \cite{lehman2011abandoning,mouret2015illuminating,pugh2016quality,cully2015robots,stanley2017openendedness}. These criteria adapt as feedback accumulates, while human judgement and expert oversight can steer selection toward scientifically promising regions that lack immediate computational validation \cite{settles2009active,gelbart2014bayesian,kandasamy2017multi}. Consequently, selection in EI functions not only to filter candidates, but also to steer the search trajectory toward evolving scientific priorities.

% 反馈源自何处。 在经典的 EC 中，评估反馈通常源自预定义的标量适应度函数或受控的模拟环境 \cite{holland1975adaptation,goldberg1989genetic,back1996evolutionary}。然而在科学发现中，反馈环境扩展为一个丰富的多模态谱系。该谱系涵盖了计算代理（如预测模型与不确定性估计 \cite{butler2018machine,jumper2021highly,merchant2023scaling,stokes2020deep}）、源自实验室实验的物理约束 \cite{burger2020mobile,macleod2020self,boiko2023autonomous,granda2018controlling,haese2019olympus}，以及由人类驱动的可重复性检查与专家判断 \cite{wang2023scientific,messeri2024artificial,hey2009fourth}。关键在于，这些多维信号的作用远不止于对候选对象进行排序。它们主动决定了底层模型、搜索约束和实验优先级如何为后续的发现周期进行动态校准。因此，EI 中的反馈环境充当了计算搜索过程与科学探究物理现实之间的主要接口。
\textbf{Where feedback originates.} EC practices typically derive evaluation feedback from predefined, scalar fitness functions or controlled simulated environments \cite{holland1975adaptation,goldberg1989genetic,back1996evolutionary}. In scientific discovery, however, the feedback environment includes various types of signals. These range from computational proxies, such as predictive models and uncertainty estimates \cite{butler2018machine,jumper2021highly,merchant2023scaling,stokes2020deep}, to physical constraints derived from laboratory experiments \cite{burger2020mobile,macleod2020self,boiko2023autonomous,granda2018controlling,haese2019olympus}, and human-driven reproducibility checks with expert judgement \cite{wang2023scientific,messeri2024artificial,hey2009fourth}. These signals do more than merely rank candidates; they influence how models, search constraints, and experimental priorities are adjusted for subsequent discovery cycles. Consequently, the feedback environment in EI serves as a direct link between the computational search and physical experiments.

% 当进化发生时。进化操作应用的时间尺度会影响系统是局限于局部搜索还是支持系统性的累积发现。在科学人工智能系统中，进化在多个时间尺度上运行，具体取决于反馈的延迟和成本。在设计和训练阶段，进化机制塑造模型架构（[real2019]regularized，[stanley2019designing]）和搜索配置（[jaderberg2017]population，[li2017hyperband]），并在学习过程中调整提议分布（[adaptiveDEsurvey2025]，[hansen2001completely]）。在推理和实验阶段，基础模型系统可以搜索推理路径（例如 yao2023tree、madaan2023selfrefine）、算法程序（例如 romera2024mathematical、novikov2025alphaevolve）和可检验的科学假设（例如 gepa2026、oh2025discovering、boiko2023autonomous），而实验室反馈则可以改进候选批次和实验优先级（例如 macleod2020self、burger2020mobile、shields2021bayesian）。在不同的时间尺度上，存档的先验知识和自适应搜索策略能够实现跨不同任务和科学领域的知识迁移（例如 gupta2021evolutionary、pan2010survey、finn2017model、yosinski2014transferable）。因此，演化的时机不仅仅是一个调度细节；它决定了保留的经验如何影响未来的发现周期。
\textbf{When evolution occurs.} The temporal scale at which evolutionary operations are applied influences whether a system remains confined to local search or supports systematic cumulative discovery. In scientific AI systems, evolution operates at multiple time scales, depending on the latency and cost of feedback. At the design and training stages, evolutionary mechanisms shape model architectures \cite{real2019regularized,stanley2019designing} and search configurations \cite{jaderberg2017population,li2017hyperband}, and adjust proposal distributions during the learning process \cite{adaptiveDEsurvey2025,hansen2001completely}. At the inference and experimental stages, foundation-model systems can search over reasoning paths \cite{yao2023tree,madaan2023selfrefine}, algorithmic programs \cite{romera2024mathematical,novikov2025alphaevolve}, and testable scientific hypotheses \cite{gepa2026,oh2025discovering,boiko2023autonomous}, while laboratory feedback refines candidate batches and experimental priorities \cite{macleod2020self,burger2020mobile,shields2021bayesian}. Across time scales, archived priors and adaptive search strategies enable knowledge transfer across distinct tasks and scientific domains \cite{gupta2021evolutionary,pan2010survey,finn2017model,yosinski2014transferable}. Consequently, the timing of evolution is more than a scheduling detail; it determines how retained experience shapes future discovery cycles.

% 总的来说，这五个维度构成了一个框架，用于比较不同领域和技术实现的各种科学人工智能系统[1]。通过考察一个系统如何整合其不断演化的对象、变异机制和反馈环境，我们可以判断它是否仅限于孤立搜索，还是能够进行系统性的累积发现。这种视角将关注点从单个算法转移到系统如何随时间学习。因此，它为分析EI循环提供了基础，EI循环整合了这些维度，将孤立的搜索轨迹转化为科学洞见。
Collectively, these five dimensions provide a framework for comparing diverse scientific AI systems across different domains and technical implementations \cite{wang2023scientific,ai4research2025}. By examining how a system combines its evolving objects, variation mechanisms, and feedback environments, we can determine whether it is limited to isolated search or capable of systematic cumulative discovery. This perspective shifts the focus from individual algorithms to how the system learns over time. Consequently, it provides a basis for analyzing the EI cycle, which integrates these dimensions to transform isolated search trajectories into scientific insight.

\section{The evolutionary intelligence cycle: from experience retention to scientific insight}\label{sec:ei-cycle}

\begin{figure}[t]
\centering
\includegraphics[width=0.98\textwidth]{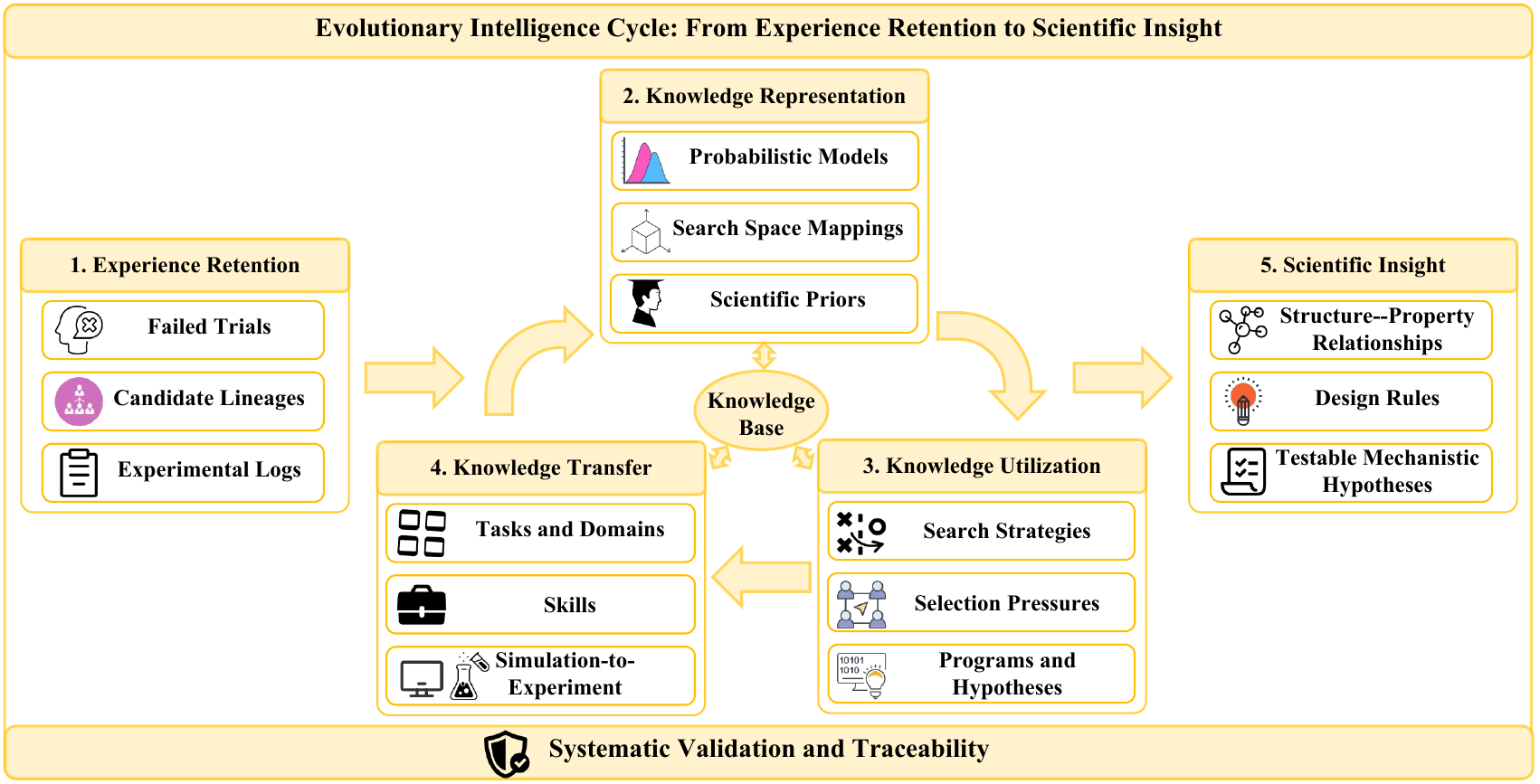}
\caption{From experience retention to scientific insight. Evolutionary intelligence (EI) transforms disconnected search trajectories into a coherent knowledge base that informs subsequent discovery cycles. \textbf{Experience retention} captures and structures diverse scientific records. \textbf{Knowledge representation} converts these raw records into interpretable representations. \textbf{Knowledge utilization} actively employs this retained memory to adjust search strategies, selection pressures, programs, and hypotheses, reshaping how candidates are generated, evaluated, and selected. \textbf{Knowledge transfer} enables this archived experience to accumulate scientific value across diverse scientific contexts. Ultimately, this cumulative process aims to produce \textbf{scientific insight}, grounded in systematic validation and traceability.}
\label{fig:ei-cycle}
\end{figure}

% 搜索轨迹的作用。EI 扩展了搜索轨迹的价值。许多 EC 应用主要关注于识别一组高性能候选对象或帕累托前沿 [deb2002fast, mouret2015illuminating, cully2015robots, pugh2016quality]。然而，对于累积发现而言，整个搜索轨迹可以作为产生科学见解的宝贵基础。EI 系统并非仅仅记录最终结果，而是经常捕获详细的证据。失败的试验往往揭示了物理和操作方面的限制，而候选对象的谱系则记录了维持或降低功能的具体修改。此外，实验记录还可以突出计算模型与物理现实之间的系统性差异 [jin2021data, macleod2020self, burger2020mobile, kudela2022critical]。通过存档这些完整的搜索历史记录，EI 将孤立的搜索轨迹转化为结构化的经验，从而为后续的发现周期提供信息。
\textbf{The role of search trajectories.} EI expands the value of search trajectories. Many EC applications primarily focus on identifying a set of high-performing candidates or a Pareto front \cite{deb2002fast,mouret2015illuminating,cully2015robots,pugh2016quality}. For cumulative discovery, however, the entire search trajectory can serve as a valuable foundation for generating scientific insight. Rather than merely recording final outcomes, EI systems frequently capture detailed evidence. Failed trials often reveal physical and operational constraints, while candidate lineages document specific modifications that preserve or degrade function. Furthermore, experimental records can highlight systematic discrepancies between computational surrogates and physical reality \cite{jin2021data,macleod2020self,burger2020mobile,kudela2022critical}. By archiving these complete search histories, EI translates isolated search trajectories into a structured experience that informs subsequent discovery cycles. This transformation is governed by the EI cycle (Fig.~\ref{fig:ei-cycle}), which systematically retains, represents, utilizes, and transfers experience to generate scientific insight.

% 从搜索保留到科学洞见。 演化智能（EI）将互不关联的搜索轨迹转化为连贯的知识库，从而为后续的发现周期提供指导。经验保留捕获并结构化多样化的科学记录，包括失败试验、候选解谱系和实验日志。知识表示将这些原始记录转化为可解释的表征，如概率代理模型、搜索空间映射和特定领域的科学先验。知识利用主动运用这些保留的记忆来同时调整搜索策略和选择压力，重塑候选解的生成、评估和选择方式。知识迁移使存档的经验能够在多样化的科学语境（包括不同的任务、领域、长度尺度以及模拟到实验的边界）中积累科学价值。最终，这一累积过程旨在产生科学洞见——表现为结构-性质关系、设计原则或可测试的机制假设——并建立在系统的验证和可追溯性之上。

% 经验保留。这种机制不同于标准的进化论实践，后者主要存档精英解决方案或非主导前沿[1975年 Holland 适应, 1989年 Goldberg 遗传, 2011年 Lehman 放弃, 2015年 Mouret 启发]。在科学发现中，经验保留系统通常采用更广泛的存档协议，记录搜索过程的更多方面。这些协议并非丢弃次优结果，而是经常纳入失败的试验、中间进化谱系、原始实验日志以及带有上下文元数据的人工注释[1975年 MacLeod 自我, 2020年 Burger 移动, 2009年 King 自动化, 2018年 Granda 控制]。通过将这些不同的记录构建成统一的历史记录，经验保留有助于保存更广泛的科学证据。这种广泛的保留为后续的知识表示提供了宝贵的原始素材，将孤立的个体数据点转化为连贯且可追溯的科学记录。
\textbf{Experience retention.} This mechanism differs from common EC practices of archiving primarily elite solutions or non-dominated fronts \cite{holland1975adaptation,goldberg1989genetic,lehman2011abandoning,mouret2015illuminating}. In scientific discovery, EI systems often employ broader archival protocols that record more aspects of the search process. Rather than discarding suboptimal outcomes, these protocols frequently incorporate failed trials, intermediate evolutionary lineages, and raw experimental logs, and human annotations with contextual metadata \cite{macleod2020self,burger2020mobile,king2009automation,granda2018controlling}. By structuring these diverse records into a unified historical record, EI helps preserve a broader range of scientific evidence. This extensive retention provides valuable raw material for subsequent knowledge representation, converting isolated data points into a coherent and traceable scientific record.

% 知识表示。原始档案记录通常需要转换为可用且可解释的形式，以指导未来的探索。在此阶段，EI 系统将累积的搜索历史转换为结构化表示。这些表示涵盖了从数据驱动的代理模型（将候选特征映射到性能和不确定性）[Jin 2021Data, Rasmussen 2006Gaussian, Snoek 2012Practical]到搜索空间的拓扑映射（揭示隐藏的吸引盆地）[Cully 2015Robots, Mouret 2015Illuminating]，以及源自物理定律或专家知识的领域特定科学先验[Butler 2018Machine, Wang 2023Scientific, Hey 2009Fourth]。此步骤将底层数据与高层推理连接起来。通过记录成功和失败，系统构建了一个结构化的历史记录，引导未来的搜索找到可行的选项，并防止重复犯错。
\textbf{Knowledge representation.} Raw archival records often require transformation into usable and interpretable forms to guide future exploration. In this phase, EI systems convert accumulated search history into structured representations. These representations range from data-driven probabilistic surrogate models that map candidate features to evaluation indicators \cite{jin2021data,rasmussen2006gaussian,snoek2012practical}, to topological mappings of the search space that reveal hidden basins of attraction \cite{cully2015robots,mouret2015illuminating}, and domain-specific scientific priors derived from physical laws or expert knowledge \cite{butler2018machine,wang2023scientific,hey2009fourth}. This step connects low-level data to higher-level reasoning. By recording both successes and failures, the system builds a structured history that guides future searches toward viable options and prevents repeated mistakes.

% 知识利用。这种机制利用保留的经验来指导持续的搜索过程。与依赖静态规则不同，进化智能系统会利用前一阶段构建的结构化表示来调整其搜索策略。例如，数据驱动的替代模型和不确定性估计可以指导新候选者的生成，引导进化变异向搜索空间中具有前景或不确定性的区域发展[1]。拓扑映射和新颖性指标会改变选择压力，惩罚过早收敛，并鼓励探索多样化的行为[1]。此外，基础模型利用编码的科学先验知识来提出方案和有效假设，从而减少对随机突变的依赖\cite{romera2024mathematical,fernando2023promptbreeder,boiko2023autonomous,gepa2026}。通过这些机制，知识利用将搜索从试错法转变为证据驱动的探究，在开放式探索和有针对性的改进之间取得平衡，并随着新反馈的到来不断改进。
\textbf{Knowledge utilization.} This mechanism uses retained experience to steer the ongoing search process. Rather than relying on static rules, EI systems adjust their search strategies using the structured representations built in the previous phase. For instance, data-driven surrogates and uncertainty estimates can guide the generation of new candidates, directing evolutionary variation toward regions of the search space that are promising or uncertain \cite{jin2021data,snoek2012practical,shahriari2016taking}. Topological mappings and novelty metrics change selection pressures, penalizing premature convergence and encouraging the exploration of diverse behaviors \cite{lehman2011abandoning,mouret2015illuminating,pugh2016quality}. Furthermore, foundation models use encoded scientific priors to propose programs and valid hypotheses, reducing reliance on random mutation \cite{romera2024mathematical,fernando2023promptbreeder,boiko2023autonomous,gepa2026}. Through these mechanisms, knowledge utilization shifts the search from trial-and-error to evidence-driven inquiry, balancing open-ended exploration and targeted refinement as new feedback arrives.

% 知识迁移。这种机制能够通过跨任务重用经验来实现累积发现，而不是将搜索限制在单次运行或密切相关的问题上[参考文献：brest2006self,gupta2021evolutionary,gupta2016multifactorial]。在现代进化智能系统中，结构化记录和自适应策略可以在新的科学背景下重用。这些系统并非从零开始，而是利用已存档的先验知识来加速在不同任务和科学领域的探索[参考文献：wang2023scientific,ai4research2025,boiko2023autonomous,agentgym2024]。这种迁移也弥合了计算和物理之间的界限。例如，在高通量模拟中学习到的启发式方法和数据驱动的代理模型可以迁移到指导成本高昂的物理实验中，从而有助于将模拟与物理实验联系起来[参考文献：jin2021data,wang2022solving,yosinski2014transferable]。类似地，实验方案和设计原则可以跨越不同的长度尺度或材料类别进行传播[参考文献：merchant2023scaling,batatia2022mace,butler2018machine]。这种知识转移的另一个关键方面是模块化技能的提取和重用。当系统发现有效的算法或稳健的提示模板时，它们会将这些逻辑结构存档为可重用的技能[参考文献：voyager2023,fernando2023promptbreeder,gepa2026,skillrl2026]。通过重新组合这些经过验证的模块，系统可以构建新的解决方案，并创建一个不断增长的科学启发式库。通过在不同的背景下不断重用历史证据，知识转移使得每个发现周期都能建立在不断增长的科学知识体系之上。
\textbf{Knowledge transfer.} This mechanism enables cumulative discovery by reusing experience across tasks, rather than limiting search to single runs or closely related problems \cite{brest2006self,gupta2021evolutionary,gupta2016multifactorial}. In EI systems, structured records and adaptive strategies are reused across new scientific contexts. Rather than starting from scratch, these systems use archived priors to accelerate exploration in distinct tasks and scientific domains \cite{wang2023scientific,ai4research2025,boiko2023autonomous,agentgym2024}. This transfer also bridges computational and physical boundaries. For instance, heuristics and data-driven models learned in simulations can be transferred to guide costly physical experiments, helping to connect simulations with physical experiments \cite{jin2021data,wang2022solving,yosinski2014transferable}. Similarly, experimental protocols and design principles can propagate across different length scales or material classes \cite{merchant2023scaling,batatia2022mace,butler2018machine}. Systems also extract and reuse modular skills. When they find effective algorithms or prompt templates, they archive these structures for future tasks \cite{voyager2023,fernando2023promptbreeder,gepa2026,skillrl2026}. Systems can then combine these saved skills to solve new problems. By reusing past results, each discovery cycle builds directly on previous knowledge rather than starting from scratch.

% 科学洞见。这一阶段将重点从寻找高性能候选者转移到生成可解释的科学知识。虽然许多搜索过程最终只找到一个最佳候选者或帕累托前沿，但累积发现系统会从整个搜索轨迹中提取出普遍适用的原则。这些洞见包括明确的结构-属性关系、设计规则或可检验的机制假设，它们解释了为什么某些候选者成功而另一些失败[wang2023scientific,hey2009fourth,messeri2024artificial]。由于这些洞见基于已存档的失败案例、多样化的搜索路径和物理验证，因此它们具有高度的可靠性和可追溯性[kudela2022critical,macleod2020self]。随着时间的推移，这些新的科学洞见会反馈到更广泛的研究群体中，增进人类的理解，并为未来的自主发现系统提供初始先验信息。
\textbf{Scientific insight.} This phase shifts the focus from finding high-performing candidates to generating interpretable scientific knowledge. While many search processes conclude with a single best candidate or a Pareto front, cumulative discovery systems extract generalized principles from the entire search trajectory. These insights include explicit structure--property relationships, design rules, or testable mechanistic hypotheses that explain why certain candidates succeed while others fail \cite{wang2023scientific,hey2009fourth,messeri2024artificial}. Because these insights are based on archived failures, diverse search paths, and physical validation, they are highly reliable and traceable \cite{kudela2022critical,macleod2020self}. Over time, this new scientific insight feeds back into the broader research community, improving human understanding and providing initial priors for future autonomous discovery systems.

\section{Modes of EI-enabled scientific discovery}\label{sec:EISD}
% 为了展示 EI 框架的普适性，我们通过演化对象而非特定应用领域的视角来分析科学 AI 系统。如图 4 所示，这一视角揭示了演化范围的递进：从演化具体的科学实体，到修改预测模型与表征，进而演化符号推理过程，并最终达到编排自动化研究工作流。这种结构性分类法阐明，尽管特定的候选空间与反馈机制在各学科间存在差异，但赋能累积式发现的底层 EI 架构保持一致 \cite{wang2023scientific,ai4research2025}。 As illustrated in Fig.~\ref{fig:modes}, t
To demonstrate the generality of the EI framework, we analyze scientific AI systems based on their evolving object rather than their specific application domains. This approach shows a progression in the scope of evolution: from concrete scientific entities and computational models, through symbolic reasoning processes, to automated research workflows. This classification shows that while specific candidate spaces and feedback mechanisms vary across disciplines, the EI architecture remains consistent \cite{wang2023scientific,ai4research2025}.

% 演化具体的科学实体。演化实体（EI）作用于物理或化学目标，包括分子（jumper，2021highly）、蛋白质（watson，2023broadly，dauparas，2022robust）和材料（merchant，2023scaling）。由于这些领域的实验反馈通常成本高昂且稀少，系统通常依赖数据驱动的替代模型来指导探索（jin，2021data）。在这种情况下，演化实体系统通常会存档失败的合成路线和负面表征数据。系统不会丢弃这些结果，而是利用它们来识别物理上可行的区域。这种保留的经验使得后续的搜索周期能够避开无法访问的区域，并专注于可行的空间，正如自主材料平台（chu，2026programmable，ghareeb，2026multi，gottweis，2026accelerating）所展示的那样。
\textbf{Evolving concrete scientific entities.} EI operates on physical or chemical targets, including molecules \cite{jumper2021highly}, proteins \cite{watson2023broadly,dauparas2022robust}, and materials \cite{merchant2023scaling}. Because experimental feedback in these domains is often costly and sparse, systems frequently rely on data-driven surrogate models to guide exploration \cite{jin2021data}. EI systems in this context often archive failed synthesis routes and negative characterization data. Rather than discarding these outcomes, the system uses them to identify physically feasible regions. This retained experience allows subsequent search cycles to avoid inaccessible areas and focus on viable spaces, as demonstrated in autonomous materials platforms \cite{chu2026programmable,ghareeb2026multi,gottweis2026accelerating}.

% 修改计算模型和表征。EI 操作于用于对科学实体进行建模的计算工具。这些系统并非直接搜索物理目标，而是演化出神经架构（[real] 和 [stanley] 分别用于正则化和设计）、分子表征（[xie] 和 [crystal] 分别用于晶体和分子表征）以及数据驱动的代理模型（[jin] 和 [data] 分别用于数据）。调整这些表征可以重塑搜索空间，使先前无法访问的区域更容易探索。在此层面上，系统会记录不同的表征如何捕捉物理或化学约束。由于这种经验可以迁移到相关任务中，因此系统能够选择有利于物理上合理解决方案的表征。例如，将基础模型集成到蛋白质和材料设计中，使用预训练的表示来指导搜索可行的生物或化学结构\cite{abramson2024alphafold3,rives2021biological,lin2023evolutionary,baek2021accurate,akiba2025evolutionary}。
\textbf{Modifying computational models and representations.} EI operates on the computational tools used to model scientific entities. Rather than searching directly over physical targets, these systems evolve neural architectures \cite{real2019regularized,stanley2019designing} and molecular representations \cite{xie2018crystal,schutt2018schnet,batzner2022e3,batatia2022mace,ward2016general}, and data-driven surrogate models \cite{jin2021data}. Adapting these representations can reshape the search space, making previously inaccessible regions easier to explore. At this level, the system records how different representations capture physical or chemical constraints. Because this experience can transfer across related tasks, it enables the system to select representations that favor physically plausible solutions. For example, integrating foundation models into protein and materials design uses pretrained representations to guide the search toward viable biological or chemical structures \cite{abramson2024alphafold3,rives2021biological,lin2023evolutionary,baek2021accurate,akiba2025evolutionary}.

% 演化符号推理过程。在更高层面上，演化符号推理（EI）系统针对符号结构，包括可执行代码和数学表达式[romera2024mathematical,novikov2025alphaevolve]，以及机制假设。与物理实验不同，这些符号实体的反馈通常是确定性的，并且可以自动验证，通常依赖于编译器、自动定理证明器或单元测试。这种自动验证使得系统无需物理材料或仪器时间即可评估大量候选方案。演化符号推理系统不会丢弃未通过测试套件的代码，而是通常会将这些错误存档。通过分析这些失败，系统可以识别重复出现的逻辑错误，或提取部分正确的子程序以指导后续修改[madaan2023selfrefine,lehman2024evolution]。在连续的循环中，这些存档的代码组件和更正使系统能够构建更复杂的算法或数学证明，通常会产生人类可读的解决方案\cite{gepa2026,oh2025discovering}。
\textbf{Evolving symbolic reasoning processes.} At a higher level, EI systems target symbolic structures, including executable code, mathematical expressions \cite{romera2024mathematical,novikov2025alphaevolve}, and mechanistic hypotheses. Unlike physical experiments, the feedback for these symbolic entities can be automatically verified, often relying on massive parallel computing. This automated verification allows systems to evaluate a large volume of candidates. Instead of discarding code that fails a test suite, EI systems often archive these errors. By analyzing these failures, the system can identify recurring logical mistakes or extract partially correct subroutines to guide subsequent modifications \cite{madaan2023selfrefine,lehman2024evolution}. Over successive cycles, these archived code components and corrections enable the system to construct more complex algorithms or mathematical proofs, often resulting in human-readable solutions \cite{gepa2026,oh2025discovering}.

% 编排自动化研究工作流程。除了修改单个组件外，EI 系统还可以针对整个研究工作流程。这些工作流程整合了从规划到实验执行的多个科学操作[参考文献：boiko2023autonomous,agentgym2024]。在这个层面上，不断演化的对象是这些任务的执行计划。反馈通常评估整个工作流程是否成功产生最终结果，例如验证假设。由于这需要完成所有步骤，因此反馈通常比组件级信号更稀疏、更延迟[参考文献：dgm2026,gottweis2026accelerating]。为了管理这些流程，EI 系统会存档成功的多步骤协议。通过保留这些执行计划，系统可以重用它们来自动化复杂的科学任务[参考文献：voyager2023,sage2025,agent02025,skillrl2026]。
\textbf{Orchestrating automated research workflows.} Beyond modifying single components, EI systems can target entire research workflows. These workflows integrate multiple scientific operations, from planning to experimental execution \cite{boiko2023autonomous,agentgym2024}. At this level, the evolving objects are the execution plans for these tasks. Feedback typically evaluates whether the complete workflow successfully produces an outcome, such as a verified hypothesis. Because this requires completing all steps, the feedback is often sparser and more delayed than component-level signals \cite{dgm2026,gottweis2026accelerating}. To manage these processes, EI systems archive successful multi-step protocols. By retaining these execution plans, the system can reuse them to automate complex scientific tasks \cite{voyager2023,sage2025,agent02025,skillrl2026}.

% 在实践中，单一的科学工作流程通常会融合多种模式。例如，一个自动化系统可能会在合成物理材料的同时，不断改进用于分析数据的代码。无论修改的对象是什么，其核心机制始终不变：系统会将过往结果存档，以指导后续步骤。随着智能工程系统实现整个研究流程的自动化，评估方式也必须随之改变。评估不应仅仅局限于衡量最终输出的质量，而应关注系统如何验证其发现，以及如何在整个过程中保留经验。

In practice, a single scientific workflow often combines multiple modes. For instance, an automated system might evolve the code used to analyze data while simultaneously synthesizing physical materials. Regardless of the objects being modified, the core mechanism remains the same: the system archives past results to guide subsequent steps. As EI systems integrate these modes into complex workflows, evaluation must look beyond the final output. Rather than measuring only the quality of a single result, assessments need to evaluate the reliability of the entire process and the system's capacity to accumulate experience over time.

% \begin{figure}[t]
% \centering
% \includegraphics[width=0.98\textwidth]{fig4.png}
% \caption{Modes of EI-enabled scientific discovery. We categorize scientific AI systems by their evolving object, showing a progression in the scope of evolution: from concrete scientific entities and computational models, through symbolic reasoning processes, to automated research workflows. In practice, these modes often overlap within a single workflow. Despite domain-specific variations in candidate spaces and feedback mechanisms, the EI architecture remains consistent across these modes.}
% \label{fig:modes}
% \end{figure}
% % \caption{基于智能的科学发现模式。我们根据科学人工智能系统所面向的演化对象对其进行分类，展示了演化范围的逐步扩展：从具体的科学实体和计算模型，到符号推理过程，再到自动化研究工作流程。在实践中，这些模式经常在单个工作流程中重叠。尽管候选空间和反馈机制存在领域差异，但底层智能架构在这些模式中保持一致。}

\section{Evaluation, challenges and roadmap}\label{sec:evaluation}

% 评估用于科学发现的EI（智能工程）需要超越单个候选模型性能的评价标准。虽然候选模型的质量是必要的基准，但EI的核心能力在于其系统性地保留经验、促进可靠的知识转移以及在迭代搜索周期中产生可解释的科学见解的能力。提升这些累积能力会带来一系列相互关联的挑战。这些挑战包括：建立评估指标并确保单个系统内部流程的可追溯性；弥合模拟与实验反馈之间的差距；以及构建可互​​操作的基础设施以进行更广泛的经验共享。
Evaluating EI for scientific discovery requires criteria that go beyond the performance of isolated candidates. While candidate quality is a necessary baseline, EI is distinguished by its ability to retain experience, transfer knowledge, and produce interpretable insights across evolutionary cycles. Achieving this cumulative discovery presents several key challenges: defining evaluation metrics, improving process traceability, bridging simulation-to-experiment gaps, and developing shared databases for experience reuse.

% 评估超越候选优化的累积发现。目前的评估框架主要量化孤立发现周期内最终候选者的质量。这种横截面指标通常不足以区分系统是积累了可迁移的知识，还是仅仅过度拟合了特定任务[1]。因此，这种局限性使得区分瞬时候选优化和系统性累积发现变得困难。弥补这一差距需要转向纵向指标。这些指标不应依赖于单次运行的性能，而应跟踪连续周期中的知识利用率和迁移效率，以及可解释的科学见解的出现[1]。
\textbf{Evaluating cumulative discovery beyond candidate refinement.} Current evaluation frameworks primarily measure the quality of final candidates within isolated discovery cycles. Such single-cycle metrics are often insufficient to distinguish whether a system accumulates transferable knowledge or merely overfits to a specific task \cite{wang2023scientific,messeri2024artificial}. Consequently, this limitation makes it difficult to separate short-term candidate refinement from systematic cumulative discovery. Addressing this gap requires a shift toward longitudinal metrics. Rather than relying on single-run performance, these metrics need to track how knowledge is utilized and transferred across successive cycles, as well as the emergence of interpretable scientific insights \cite{krenn2022predicting,boiko2023autonomous}.

% 演化过程的可追溯性。 当基础模型生成变异时，追踪特定候选对象是如何推导出来的通常很困难。如果没有清晰的生成过程记录，科学家就无法独立验证或重现结果 \cite{lu2026automation,aygun2026ai}。为了解决这个问题，系统必须记录候选修改的完整历史，包括成功的步骤和失败的尝试。此外，让人类专家审查这些记录有助于确保生成的候选符合已知的物理或逻辑规则，从而将原始的模型输出转化为可信的科学证据 \cite{madaan2023selfrefine,messeri2024artificial}。

\textbf{Traceability of evolutionary processes.} When foundation models generate variations, tracing how a specific candidate was derived is often difficult. Without a clear record of the evolutionary process, scientists cannot independently verify or reproduce the results \cite{lu2026automation,aygun2026ai}. To address this, systems must record the complete history of candidate modifications, including both successful steps and failed attempts. Additionally, involving human experts to review these records helps ensure that the generated candidates align with known physical or logical rules, turning raw model outputs into trustworthy scientific evidence \cite{madaan2023selfrefine,messeri2024artificial}.

% 反馈可靠性与模拟到实验的差异。 EI 循环高度依赖反馈，而反馈往往成本高昂、稀疏且带有噪声。真实世界的实验缓慢且昂贵，而计算模拟虽然快速，但通常会简化物理定律 \cite{shields2021bayesian,burger2020mobile}。仅仅依赖其中任何一种来源都可能误导搜索。为了解决这个问题，系统必须结合模拟与实验。它们需要测量模拟预测与真实实验结果之间的差距，并利用这种差异来不断修正其模型，从而避免进入物理上不可能的区域 \cite{kandasamy2017multi,gelbart2014bayesian}。

\textbf{Feedback reliability and simulation-to-experiment gaps.} The EI cycle depends heavily on feedback, which is often costly, sparse, and noisy. Real-world experiments are slow and expensive, while computational simulations are fast but frequently simplify physical laws \cite{shields2021bayesian,burger2020mobile}. Relying solely on either source can mislead the search. To address this, systems must combine simulations with experiments. They need to measure the gap between simulated predictions and real experimental outcomes, using this difference to continuously correct their models and avoid physically impossible regions \cite{kandasamy2017multi,gelbart2014bayesian}.

% 用于经验迁移的互操作基础设施。 目前许多 EI 系统是独立运行的，它们产生了大量其他研究者难以复用的数据。缺乏共享的数据平台，阻碍了知识在不同科学领域、长度尺度以及模拟到实验边界之间的迁移 \cite{butler2018machine,merchant2023scaling}。为了解决这个问题，科学界需要统一的数据平台，来系统地记录失败的实验、修改历史和训练好的模型。制定通用的数据格式与共享标准，将使新的 EI 系统能够利用先前积累的经验来启动搜索，从而推动更广泛研究社区的累积式科学发现 \cite{ai4research2025,wang2023scientific}。
\textbf{Shared infrastructure for experience reuse.} Many current EI systems operate independently, generating large amounts of data that are difficult for other researchers to reuse. The lack of shared data platforms hinders the transfer of knowledge across different scientific domains, length scales, and simulation-to-experiment boundaries \cite{butler2018machine,merchant2023scaling}. To address this, the community needs standardized databases that systematically record failed experiments, modification histories, and trained models. Developing common data formats and sharing standards would allow new EI systems to begin their searches using previously accumulated experience, enabling cumulative scientific discovery across the broader research community \cite{ai4research2025,wang2023scientific}.

\section{Outlook}\label{sec:outlook}

% 向面向科学发现的 EI 的转变，将重点从自动化单一任务转移到了随时间积累知识。未来的科学 AI 系统可能会结合基础模型与自动化实验，并在演化循环的引导下运行 \cite{wang2023scientific,ai4research2025}。在这些系统中，每一个被评估的候选与失败的实验都会增加到一个共享的记录中。这种历史使得后续的搜索能够利用现有的科学知识开始，而不是从零开始。

The transition to EI for scientific discovery shifts the focus from automating single tasks to accumulating knowledge over time. Future scientific AI systems will likely integrate foundation models with automated experiments through evolutionary cycles \cite{wang2023scientific,ai4research2025}. In these systems, every evaluated candidate and failed experiment contributes to a shared record. This accumulated history allows subsequent searches to build on existing scientific knowledge rather than starting from scratch.

% 随着这些系统变得更加自主，人类研究人员将把更多精力放在设定总体目标和验证结果上。科学家在应用领域知识和解释出乎意料的实验结果方面依然不可或缺 \cite{messeri2024artificial,lu2026automation}。因此，EI 的成功不应仅仅通过计算速度或生成的候选数量来衡量。相反，应通过其产生清晰、有实验支持的洞见的能力来评判，这些洞见能够经受住科学审查，并帮助研究人员理解底层的科学。

As these systems become more autonomous, the role of human researchers will shift toward setting overall goals and validating results. Scientists remain essential for applying domain expertise and interpreting unexpected experimental outcomes \cite{messeri2024artificial,lu2026automation}. Therefore, the success of EI should not be judged solely by computational speed, the number of candidates generated, and accuracy. Instead, it depends on the system's ability to produce experimentally validated insights that withstand scientific scrutiny and help researchers understand the underlying mechanisms.

% \section*{Acknowledgements}

% \section*{Author contributions}

% \section*{Competing interests}

\bibliography{ei_for_science_sn_bibliography}

@article{ai4research2025,
  title={Ai4research: A survey of artificial intelligence for scientific research},
  author={Chen, Qiguang and Yang, Mingda and Qin, Libo and Liu, Jinhao and Yan, Zheng and Guan, Jiannan and Peng, Dengyun and Ji, Yiyan and Li, Hanjing and Hu, Mengkang and others},
  journal={arXiv preprint arXiv:2507.01903},
  year={2025}
}

@article{zhang2026evom,
  title={EVOM: Agentic Meta-Evolution of Actor-Critic Architectures for Reinforcement Learning},
  author={Zhang, Boyun and Wang, Chao and Wu, Kai},
  journal={arXiv preprint arXiv:2606.26327},
  year={2026}
}

@article{eiben2015evolution,
  title={From evolutionary computation to the evolution of things},
  author={Eiben, Agoston E and Smith, Jim},
  journal={Nature},
  volume={521},
  number={7553},
  pages={476--482},
  year={2015},
  publisher={Nature Publishing Group UK London}
}

@article{cussatblanc2020biological,
  title={A biological perspective on evolutionary computation},
  author={Miikkulainen, Risto and Forrest, Stephanie},
  journal={Nature Machine Intelligence},
  volume={3},
  number={1},
  pages={9--15},
  year={2021},
  publisher={Nature Publishing Group UK London}
}

@article{agentic_science2025,
  title={From ai for science to agentic science: A survey on autonomous scientific discovery},
  author={Wei, Jiaqi and Yang, Yuejin and Zhang, Xiang and Chen, Yuhan and Zhuang, Xiang and Gao, Zhangyang and Zhou, Dongzhan and Wang, Guangshuai and Gao, Zhiqiang and Cao, Juntai and others},
  journal={arXiv preprint arXiv:2508.14111},
  year={2025}
}

@article{scientific_intelligence2025,
  title={Towards scientific intelligence: A survey of llm-based scientific agents},
  author={Ren, Shuo and Xie, Can and Jian, Pu and Ren, Zhenjiang and Leng, Chunlin and Zhang, Jiajun},
  journal={arXiv preprint arXiv:2503.24047},
  year={2025}
}

@article{arxiv2510_14150,
  title={Codeevolve: An open source evolutionary coding agent for algorithm discovery and optimization},
  author={Assump{\c{c}}{\~a}o, Henrique and Ferreira, Diego and Campos, Leandro and Murai, Fabricio},
  journal={arXiv preprint arXiv:2510.14150},
  year={2025}
}

@article{arxiv2410_05080,
      title={ScienceAgentBench: Toward Rigorous Assessment of Language Agents for Data-Driven Scientific Discovery}, 
      author={Ziru Chen and Shijie Chen and Yuting Ning and Qianheng Zhang and Boshi Wang and Botao Yu and Yifei Li and Zeyi Liao and Chen Wei and Zitong Lu and Vishal Dey and Mingyi Xue and Frazier N. Baker and Benjamin Burns and Daniel Adu-Ampratwum and Xuhui Huang and Xia Ning and Song Gao and Yu Su and Huan Sun},
      year={2024},
      journal={arXiv preprint arXiv:2410.05080}
}

@article{shahriari2016taking,
  title={Taking the human out of the loop: A review of Bayesian optimization},
  author={Shahriari, Bobak and Swersky, Kevin and Wang, Ziyu and Adams, Ryan P and De Freitas, Nando},
  journal={Proceedings of the IEEE},
  volume={104},
  number={1},
  pages={148--175},
  year={2015},
  publisher={IEEE}
}

@article{frazier2018tutorial,
  title={A tutorial on Bayesian optimization},
  author={Frazier, Peter I},
  journal={arXiv preprint arXiv:1807.02811},
  year={2018}
}

@article{shields2021bayesian,
  title={Bayesian reaction optimization as a tool for chemical synthesis},
  author={Shields, Benjamin J and Stevens, Jason and Li, Jun and Parasram, Marvin and Damani, Farhan and Alvarado, Jesus I Martinez and Janey, Jacob M and Adams, Ryan P and Doyle, Abigail G},
  journal={Nature},
  volume={590},
  number={7844},
  pages={89--96},
  year={2021},
  publisher={Nature Publishing Group UK London}
}

@article{wang2023scientific,
  title={Scientific discovery in the age of artificial intelligence},
  author={Wang, Hanchen and Fu, Tianfan and Du, Yuanqi and Gao, Wenhao and Huang, Kexin and Liu, Ziming and Chandak, Payal and Liu, Shengchao and Van Katwyk, Peter and Deac, Andreea and others},
  journal={Nature},
  volume={620},
  number={7972},
  pages={47--60},
  year={2023},
  publisher={Nature Publishing Group UK London}
}

@article{king2009automation,
  title={The automation of science},
  author={King, Ross D and Rowland, Jem and Oliver, Stephen G and Young, Michael and Aubrey, Wayne and Byrne, Emma and Liakata, Maria and Markham, Magdalena and Pir, Pinar and Soldatova, Larisa N and others},
  journal={Science},
  volume={324},
  number={5923},
  pages={85--89},
  year={2009},
  publisher={American Association for the Advancement of Science}
}

@article{butler2018machine,
  title={Machine learning for molecular and materials science},
  author={Butler, Keith T and Davies, Daniel W and Cartwright, Hugh and Isayev, Olexandr and Walsh, Aron},
  journal={Nature},
  volume={559},
  number={7715},
  pages={547--555},
  year={2018},
  publisher={Nature Publishing Group UK London}
}

@techreport{settles2009active,
  author = {Settles, Burr},
  title = {Active Learning Literature Survey},
  institution = {University of Wisconsin--Madison},
  year = {2009}
}

@article{macleod2020self,
  title={Self-driving laboratory for accelerated discovery of thin-film materials},
  author={MacLeod, Benjamin P and Parlane, Fraser GL and Morrissey, Thomas D and H{\"a}se, Florian and Roch, Lo{\"\i}c M and Dettelbach, Kevan E and Moreira, Raphaell and Yunker, Lars PE and Rooney, Michael B and Deeth, Joseph R and others},
  journal={Science Advances},
  volume={6},
  number={20},
  pages={eaaz8867},
  year={2020},
  publisher={American Association for the Advancement of Science}
}

@article{messeri2024artificial,
  title={Artificial intelligence and illusions of understanding in scientific research},
  author={Messeri, Lisa and Crockett, Molly J},
  journal={Nature},
  volume={627},
  number={8002},
  pages={49--58},
  year={2024},
  publisher={Nature Publishing Group UK London}
}

@book{holland1975adaptation,
  title={Adaptation in natural and artificial systems: an introductory analysis with applications to biology, control, and artificial intelligence},
  author={Holland, John H},
  year={1992},
  publisher={MIT press}
}

@article{goldberg1989genetic,
  title={Classifier systems and genetic algorithms},
  author={Booker, Lashon B. and Goldberg, David E. and Holland, John H.},
  journal={Artificial intelligence},
  volume={40},
  number={1-3},
  pages={235--282},
  year={1989},
  publisher={Elsevier}
}

@article{koza1992genetic,
  title={Genetic programming as a means for programming computers by natural selection},
  author={Koza, John R},
  journal={Statistics and computing},
  volume={4},
  number={2},
  pages={87--112},
  year={1994},
  publisher={Springer}
}

@book{back1996evolutionary,
  title={Evolutionary algorithms in theory and practice: evolution strategies, evolutionary programming, genetic algorithms},
  author={B{\"a}ck, Thomas},
  year={1996},
  publisher={Oxford university press}
}

@article{hansen2001completely,
  title={Completely derandomized self-adaptation in evolution strategies},
  author={Hansen, Nikolaus and Ostermeier, Andreas},
  journal={Evolutionary computation},
  volume={9},
  number={2},
  pages={159--195},
  year={2001},
  publisher={MIT Press}
}

@article{wierstra2014natural,
  title={Natural evolution strategies},
  author={Wierstra, Daan and Schaul, Tom and Glasmachers, Tobias and Sun, Yi and Peters, Jan and Schmidhuber, J{\"u}rgen},
  journal={The Journal of Machine Learning Research},
  volume={15},
  number={1},
  pages={949--980},
  year={2014},
  publisher={JMLR. org}
}

@article{deb2002fast,
  title={A fast and elitist multiobjective genetic algorithm: NSGA-II},
  author={Deb, Kalyanmoy and Pratap, Amrit and Agarwal, Sameer and Meyarivan, TAMT},
  journal={IEEE transactions on evolutionary computation},
  volume={6},
  number={2},
  pages={182--197},
  year={2002},
  publisher={Ieee}
}

@article{zhang2007moead,
  title={MOEA/D: A multiobjective evolutionary algorithm based on decomposition},
  author={Zhang, Qingfu and Li, Hui},
  journal={IEEE Transactions on evolutionary computation},
  volume={11},
  number={6},
  pages={712--731},
  year={2007},
  publisher={IEEE}
}

@article{lehman2011abandoning,
  title={Abandoning objectives: Evolution through the search for novelty alone},
  author={Lehman, Joel and Stanley, Kenneth O},
  journal={Evolutionary computation},
  volume={19},
  number={2},
  pages={189--223},
  year={2011},
  publisher={MIT Press}
}

@article{mouret2015illuminating,
  title={Illuminating search spaces by mapping elites},
  author={Mouret, Jean-Baptiste and Clune, Jeff},
  journal={arXiv preprint arXiv:1504.04909},
  year={2015}
}

@article{jin2021data,
  title={Data-driven evolutionary optimization: An overview and case studies},
  author={Jin, Yaochu and Wang, Handing and Chugh, Tinkle and Guo, Dan and Miettinen, Kaisa},
  journal={IEEE Transactions on Evolutionary Computation},
  volume={23},
  number={3},
  pages={442--458},
  year={2018},
  publisher={IEEE}
}

@article{kudela2022critical,
  title={A critical problem in benchmarking and analysis of evolutionary computation methods},
  author={Kudela, Jakub},
  journal={Nature Machine Intelligence},
  volume={4},
  number={12},
  pages={1238--1245},
  year={2022},
  publisher={Nature Publishing Group UK London}
}

@article{larranaga2002estimation,
  author={Larrañaga, Pedro and Bielza, Concha},
  journal={IEEE Transactions on Evolutionary Computation}, 
  title={Estimation of Distribution Algorithms in Machine Learning: A Survey}, 
  year={2024},
  volume={28},
  number={5},
  pages={1301-1321},
  doi={10.1109/TEVC.2023.3314105}}

@article{rubinstein1999cross,
  title={The cross-entropy method for combinatorial and continuous optimization},
  author={Rubinstein, Reuven},
  journal={Methodology and computing in applied probability},
  volume={1},
  number={2},
  pages={127--190},
  year={1999},
  publisher={Springer}
}

@article{pugh2016quality,
  title={Quality diversity: A new frontier for evolutionary computation},
  author={Pugh, Justin K and Soros, Lisa B and Stanley, Kenneth O},
  journal={Frontiers in Robotics and AI},
  volume={3},
  pages={40},
  year={2016},
  publisher={Frontiers Media SA}
}

@article{stanley2002evolving,
  title={Evolving neural networks through augmenting topologies},
  author={Stanley, Kenneth O and Miikkulainen, Risto},
  journal={Evolutionary computation},
  volume={10},
  number={2},
  pages={99--127},
  year={2002},
  publisher={MIT Press}
}

@article{stanley2019designing,
  title={Designing neural networks through neuroevolution},
  author={Stanley, Kenneth O and Clune, Jeff and Lehman, Joel and Miikkulainen, Risto},
  journal={Nature Machine Intelligence},
  volume={1},
  number={1},
  pages={24--35},
  year={2019},
  publisher={Nature Publishing Group UK London}
}

@article{real2019regularized,
  title={Regularized evolution for image classifier architecture search},
  author={Real, Esteban and Aggarwal, Alok and Huang, Yanping and Le, Quoc V},
  journal={Proceedings of the aaai conference on artificial intelligence},
  volume={33},
  number={01},
  pages={4780--4789},
  year={2019}
}

@article{jaderberg2017population,
  title={Population based training of neural networks},
  author={Jaderberg, Max and Dalibard, Valentin and Osindero, Simon and Czarnecki, Wojciech M and Donahue, Jeff and Razavi, Ali and Vinyals, Oriol and Green, Tim and Dunning, Iain and Simonyan, Karen and others},
  journal={arXiv preprint arXiv:1711.09846},
  year={2017}
}

@article{gupta2016multifactorial,
  title={Multifactorial evolution: Toward evolutionary multitasking},
  author={Gupta, Abhishek and Ong, Yew-Soon and Feng, Liang},
  journal={IEEE Transactions on Evolutionary Computation},
  volume={20},
  number={3},
  pages={343--357},
  year={2015},
  publisher={IEEE}
}

@article{gupta2021evolutionary,
  title={Evolutionary transfer optimization-a new frontier in evolutionary computation research},
  author={Tan, Kay Chen and Feng, Liang and Jiang, Min},
  journal={IEEE Computational Intelligence Magazine},
  volume={16},
  number={1},
  pages={22--33},
  year={2021},
  publisher={IEEE}
}

@article{wang2022solving,
  title={Solving multitask optimization problems with adaptive knowledge transfer via anomaly detection},
  author={Wang, Chao and Liu, Jing and Wu, Kai and Wu, Zhaoyang},
  journal={IEEE Transactions on Evolutionary Computation},
  volume={26},
  number={2},
  pages={304--318},
  year={2021},
  publisher={IEEE}
}

@article{pan2010survey,
  title={A survey on transfer learning},
  author={Pan, Sinno Jialin and Yang, Qiang},
  journal={IEEE Transactions on knowledge and data engineering},
  volume={22},
  number={10},
  pages={1345--1359},
  year={2009},
  publisher={IEEE}
}

@article{adaptiveDEsurvey2025,
  title={A comprehensive survey of adaptive strategies in differential evolutionary algorithms},
  author={Ye, Xinggui and Li, Jianping and Wang, Peng and Suganthan, Ponnuthurai Nagaratnam},
  journal={Swarm and Evolutionary Computation},
  volume={98},
  pages={102081},
  year={2025},
  publisher={Elsevier}
}

@article{rasmussen2006gaussian,
  title={Gaussian processes for machine learning},
  author={Seeger, Matthias},
  journal={International journal of neural systems},
  volume={14},
  number={02},
  pages={69--106},
  year={2004},
  publisher={World Scientific}
}

@article{snoek2012practical,
  title={Practical bayesian optimization of machine learning algorithms},
  author={Snoek, Jasper and Larochelle, Hugo and Adams, Ryan P},
  journal={Advances in neural information processing systems},
  volume={25},
  year={2012}
}

@article{burger2020mobile,
  title={A mobile robotic chemist},
  author={Burger, Benjamin and Maffettone, Phillip M and Gusev, Vladimir V and Aitchison, Catherine M and Bai, Yang and Wang, Xiaoyan and Li, Xiaobo and Alston, Ben M and Li, Buyi and Clowes, Rob and others},
  journal={Nature},
  volume={583},
  number={7815},
  pages={237--241},
  year={2020},
  publisher={Nature Publishing Group UK London}
}

@article{boiko2023autonomous,
  title={Autonomous chemical research with large language models},
  author={Boiko, Daniil A and MacKnight, Robert and Kline, Ben and Gomes, Gabe},
  journal={Nature},
  volume={624},
  number={7992},
  pages={570--578},
  year={2023},
  publisher={Nature Publishing Group UK London}
}

@article{romera2024mathematical,
  title={Mathematical discoveries from program search with large language models},
  author={Romera-Paredes, Bernardino and Barekatain, Mohammadamin and Novikov, Alexander and Balog, Matej and Kumar, M Pawan and Dupont, Emilien and Ruiz, Francisco JR and Ellenberg, Jordan S and Wang, Pengming and Fawzi, Omar and others},
  journal={Nature},
  volume={625},
  number={7995},
  pages={468--475},
  year={2024},
  publisher={Nature Publishing Group UK London}
}

@article{novikov2025alphaevolve,
  title={Alphaevolve: A coding agent for scientific and algorithmic discovery},
  author={Novikov, Alexander and V{\~u}, Ng{\^a}n and Eisenberger, Marvin and Dupont, Emilien and Huang, Po-Sen and Wagner, Adam Zsolt and Shirobokov, Sergey and Kozlovskii, Borislav and Ruiz, Francisco JR and Mehrabian, Abbas and others},
  journal={arXiv preprint arXiv:2506.13131},
  year={2025}
}

@article{oh2025discovering,
  title={Discovering state-of-the-art reinforcement learning algorithms},
  author={Oh, Junhyuk and Farquhar, Gregory and Kemaev, Iurii and Calian, Dan A and Hessel, Matteo and Zintgraf, Luisa and Singh, Satinder and Van Hasselt, Hado and Silver, David},
  journal={Nature},
  volume={648},
  number={8093},
  pages={312--319},
  year={2025},
  publisher={Nature Publishing Group UK London}
}

@article{fernando2023promptbreeder,
  title={Promptbreeder: Self-referential self-improvement via prompt evolution},
  author={Fernando, Chrisantha and Banarse, Dylan and Michalewski, Henryk and Osindero, Simon and Rockt{\"a}schel, Tim},
  journal={arXiv preprint arXiv:2309.16797},
  year={2023}
}

@article{lehman2024evolution,
      title={Evolution through Large Models}, 
      author={Joel Lehman and Jonathan Gordon and Shawn Jain and Kamal Ndousse and Cathy Yeh and Kenneth O. Stanley},
      year={2022},
      journal={arXiv preprint arXiv:2206.08896},
}

@article{akiba2025evolutionary,
  title={Evolutionary optimization of model merging recipes},
  author={Akiba, Takuya and Shing, Makoto and Tang, Yujin and Sun, Qi and Ha, David},
  journal={Nature Machine Intelligence},
  volume={7},
  number={2},
  pages={195--204},
  year={2025},
  publisher={Nature Publishing Group UK London}
}

@article{ghareeb2026multi,
  title={A multi-agent system for automating scientific discovery},
  author={Ghareeb, Ali Essam and Chang, Benjamin and Mitchener, Ludovico and Yiu, Angela and Szostkiewicz, Caralyn J and Shved, Dmytro and Gyimesi, Gavin J and Laurent, Jon M and Wright, Samantha M and Razzak, Muhammed T and others},
  journal={Nature},
  pages={1--3},
  year={2026},
  publisher={Nature Publishing Group UK London}
}

@article{gottweis2026accelerating,
  title={Accelerating scientific discovery with Co-Scientist},
  author={Gottweis, Juraj and Weng, Wei-Hung and Daryin, Alexander and Tu, Tao and Sirkovic, Petar and Myaskovsky, Artiom and Glowaty, Grzegorz and Weissenberger, Felix and Orlandi, Alessio and Popovici, Dan and others},
  journal={Nature},
  pages={1--3},
  year={2026},
  publisher={Nature Publishing Group UK London}
}

@article{aygun2026ai,
  title={An AI system to help scientists write expert-level empirical software},
  author={Ayg{\"u}n, Eser and Belyaeva, Anastasiya and Comanici, Gheorghe and Coram, Marc and Cui, Hao and Garrison, Jake and Johnston, Renee and Kast, Anton and McLean, Cory Y and Norgaard, Peter and others},
  journal={Nature},
  pages={1--3},
  year={2026},
  publisher={Nature Publishing Group UK London}
}

@article{lu2026automation,
  title={Towards end-to-end automation of AI research},
  author={Lu, Chris and Lu, Cong and Lange, Robert Tjarko and Yamada, Yutaro and Hu, Shengran and Foerster, Jakob and Ha, David and Clune, Jeff},
  journal={Nature},
  volume={651},
  number={8107},
  pages={914--919},
  year={2026},
  publisher={Nature Publishing Group UK London}
}

@article{chu2026programmable,
  title={Programmable RNA translation through deep learning-driven IRES discovery and de novo generation},
  author={Chu, Yanyi and Yin, Di and Yu, Dan and Xu, Guangxue and Zhang, Junze and Wang, Xiaotong and Shen, Yue and Li, Yupeng and Zhao, Ning and Zhu, Yi and others},
  journal={Nature Machine Intelligence},
  volume={8},
  number={4},
  pages={559--574},
  year={2026},
  publisher={Nature Publishing Group UK London}
}

@article{wang2026llmEA,
  title={When large language models meet evolutionary algorithms: Potential enhancements and challenges},
  author={Wang, Chao and Zhao, Jiaxuan and Jiao, Licheng and Li, Lingling and Liu, Fang and Yang, Shuyuan},
  journal={Research},
  volume={8},
  pages={0646},
  year={2025},
  publisher={AAAS}
}

@article{selfevolvingagents2026,
  title={A Survey of Self-Evolving Agents: What, When, How, and Where to Evolve on the Path to Artificial Super Intelligence},
  author={Gao, Huan-ang and Geng, Jiayi and Hua, Wenyue and Hu, Mengkang and Juan, Xinzhe and Liu, Hongzhang and Liu, Shilong and Qiu, Jiahao and Qi, Xuan and Wu, Yiran and others},
  journal={arXiv preprint arXiv:2507.21046},
  year={2025}
}

@article{agentgym2024,
  title={Agentgym: Evolving large language model-based agents across diverse environments},
  author={Xi, Z and Ding, Y and Chen, W and Hong, B and Guo, H and Wang, J and Yang, D and Liao, C and Guo, X and He, W and others},
  journal={arXiv preprint arXiv:2406.04151},
  year={2024}
}

@article{dgm2026,
  title={Darwin godel machine: Open-ended evolution of self-improving agents},
  author={Zhang, Jenny and Hu, Shengran and Lu, Cong and Lange, Robert and Clune, Jeff},
  journal={arXiv preprint arXiv:2505.22954},
  year={2025}
}

@article{sage2025,
  title={Sage: Self-evolving agents with reflective and memory-augmented abilities},
  author={Liang, Xuechen and Tao, Meiling and Xia, Yinghui and Wang, Jianhui and Li, Kun and Wang, Yijin and He, Yangfan and Yang, Jingsong and Shi, Tianyu and Wang, Yuantao and others},
  journal={Neurocomputing},
  volume={647},
  pages={130470},
  year={2025},
  publisher={Elsevier}
}

@article{agent02025,
  title={Agent0: Unleashing self-evolving agents from zero data via tool-integrated reasoning},
  author={Xia, Peng and Zeng, Kaide and Liu, Jiaqi and Qin, Can and Wu, Fang and Zhou, Yiyang and Xiong, Caiming and Yao, Huaxiu},
  journal={arXiv preprint arXiv:2511.16043},
  year={2025}
}

@article{skillrl2026,
  title={Skillrl: Evolving agents via recursive skill-augmented reinforcement learning},
  author={Xia, Peng and Chen, Jianwen and Wang, Hanyang and Liu, Jiaqi and Zeng, Kaide and Wang, Yu and Han, Siwei and Zhou, Yiyang and Zhao, Xujiang and Chen, Haifeng and others},
  journal={arXiv preprint arXiv:2602.08234},
  year={2026}
}

@article{gepa2026,
  title={Gepa: Reflective prompt evolution can outperform reinforcement learning},
  author={Agrawal, Lakshya A and Tan, Shangyin and Soylu, Dilara and Ziems, Noah and Khare, Rishi and Opsahl-Ong, Krista and Singhvi, Arnav and Shandilya, Herumb and Ryan, Michael J and Jiang, Meng and others},
  journal={arXiv preprint arXiv:2507.19457},
  year={2025}
}

@article{sarkar2025hyperscale,
  title={Evolution strategies at the hyperscale},
  author={Sarkar, Bidipta and Fellows, Mattie and Duque, Juan Agustin and Letcher, Alistair and Villares, Antonio Le{\'o}n and Sims, Anya and Wibault, Clarisse and Samsonov, Dmitry and Cope, Dylan and Liesen, Jarek and others},
  journal={arXiv preprint arXiv:2511.16652},
  year={2025}
}

@article{voyager2023,
  title={Voyager: An open-ended embodied agent with large language models},
  author={Wang, Guanzhi and Xie, Yuqi and Jiang, Yunfan and Mandlekar, Ajay and Xiao, Chaowei and Zhu, Yuke and Fan, Linxi and Anandkumar, Anima},
  journal={arXiv preprint arXiv:2305.16291},
  year={2023}
}

@book{hey2009fourth,
  title={The fourth paradigm: data-intensive scientific discovery},
  author={Hey, Anthony JG and Tansley, Stewart and Tolle, Kristin Michele and others},
  volume={1},
  year={2009},
  publisher={Microsoft research Redmond, WA}
}

@article{fogel1966artificial,
  title={An introduction to simulated evolutionary optimization},
  author={Fogel, David B},
  journal={IEEE transactions on neural networks},
  volume={5},
  number={1},
  pages={3--14},
  year={1994},
  publisher={IEEE}
}

@article{rechenberg1973evolutionsstrategie,
  title={Evolution strategies--a comprehensive introduction},
  author={Beyer, Hans-Georg and Schwefel, Hans-Paul},
  journal={Natural computing},
  volume={1},
  number={1},
  pages={3--52},
  year={2002},
  publisher={Springer}
}

@article{salimans2017evolution,
  title={Evolution strategies as a scalable alternative to reinforcement learning},
  author={Salimans, Tim and Ho, Jonathan and Chen, Xi and Sidor, Szymon and Sutskever, Ilya},
  journal={arXiv preprint arXiv:1703.03864},
  year={2017}
}

@article{mockus1978application,
  title={The application of Bayesian methods for seeking the extremum},
  author={Mockus, Jonas},
  journal={Towards global optimization},
  volume={2},
  pages={117},
  year={1998}
}

@article{gelbart2014bayesian,
  title={Bayesian optimization with unknown constraints},
  author={Gelbart, Michael A and Snoek, Jasper and Adams, Ryan P},
  journal={arXiv preprint arXiv:1403.5607},
  year={2014}
}

@article{kandasamy2017multi,
  title={Multi-fidelity Bayesian Optimisation with Continuous Approximations},
  author={Kandasamy, Kirthevasan and Dasarathy, Gautam and Schneider, Jeff and Poczos, Barnabas},
  journal={arXiv preprint arXiv:1703.06240},
  year={2017}
}

@article{gilmer2017neural,
author = {Gilmer, Justin and Schoenholz, Samuel S. and Riley, Patrick F. and Vinyals, Oriol and Dahl, George E.},
title = {Neural message passing for Quantum chemistry},
year = {2017},
journal = {International Conference on Machine Learning},
pages = {1263–1272},
organization={PMLR}
}

@article{xie2018crystal,
  title={Crystal graph convolutional neural networks for an accurate and interpretable prediction of material properties},
  author={Xie, Tian and Grossman, Jeffrey C},
  journal={Physical review letters},
  volume={120},
  number={14},
  pages={145301},
  year={2018},
  publisher={APS}
}

@article{schutt2018schnet,
  title={Schnet--a deep learning architecture for molecules and materials},
  author={Sch{\"u}tt, Kristof T and Sauceda, Huziel E and Kindermans, P-J and Tkatchenko, Alexandre and M{\"u}ller, K-R},
  journal={The Journal of chemical physics},
  volume={148},
  number={24},
  year={2018},
  publisher={AIP Publishing}
}

@article{batzner2022e3,
  title={E (3)-equivariant graph neural networks for data-efficient and accurate interatomic potentials},
  author={Batzner, Simon and Musaelian, Albert and Sun, Lixin and Geiger, Mario and Mailoa, Jonathan P and Kornbluth, Mordechai and Molinari, Nicola and Smidt, Tess E and Kozinsky, Boris},
  journal={Nature communications},
  volume={13},
  number={1},
  pages={2453},
  year={2022},
  publisher={Nature Publishing Group UK London}
}

@article{batatia2022mace,
  title={MACE: Higher order equivariant message passing neural networks for fast and accurate force fields},
  author={Batatia, Ilyes and Kovacs, David P and Simm, Gregor and Ortner, Christoph and Cs{\'a}nyi, G{\'a}bor},
  journal={Advances in neural information processing systems},
  volume={35},
  pages={11423--11436},
  year={2022}
}

@article{ward2016general,
  title={A general-purpose machine learning framework for predicting properties of inorganic materials},
  author={Ward, Logan and Agrawal, Ankit and Choudhary, Alok and Wolverton, Christopher},
  journal={npj Computational Materials},
  volume={2},
  number={1},
  pages={16028},
  year={2016},
  publisher={Nature Publishing Group}
}

@article{jumper2021highly,
  title={Highly accurate protein structure prediction with AlphaFold},
  author={Jumper, John and Evans, Richard and Pritzel, Alexander and Green, Tim and Figurnov, Michael and Ronneberger, Olaf and Tunyasuvunakool, Kathryn and Bates, Russ and {\v{Z}}{\'\i}dek, Augustin and Potapenko, Anna and others},
  journal={nature},
  volume={596},
  number={7873},
  pages={583--589},
  year={2021},
  publisher={Nature Publishing Group UK London}
}

@article{abramson2024alphafold3,
  title={Accurate structure prediction of biomolecular interactions with AlphaFold 3},
  author={Abramson, Josh and Adler, Jonas and Dunger, Jack and Evans, Richard and Green, Tim and Pritzel, Alexander and Ronneberger, Olaf and Willmore, Lindsay and Ballard, Andrew J and Bambrick, Joshua and others},
  journal={Nature},
  volume={630},
  number={8016},
  pages={493--500},
  year={2024},
  publisher={Nature Publishing Group UK London}
}

@article{baek2021accurate,
  title={Accurate prediction of protein structures and interactions using a three-track neural network},
  author={Baek, Minkyung and DiMaio, Frank and Anishchenko, Ivan and Dauparas, Justas and Ovchinnikov, Sergey and Lee, Gyu Rie and Wang, Jue and Cong, Qian and Kinch, Lisa N and Schaeffer, R Dustin and others},
  journal={Science},
  volume={373},
  number={6557},
  pages={871--876},
  year={2021},
  publisher={American Association for the Advancement of Science}
}

@article{watson2023broadly,
  title={De novo design of protein structure and function with RFdiffusion},
  author={Watson, Joseph L and Juergens, David and Bennett, Nathaniel R and Trippe, Brian L and Yim, Jason and Eisenach, Helen E and Ahern, Woody and Borst, Andrew J and Ragotte, Robert J and Milles, Lukas F and others},
  journal={Nature},
  volume={620},
  number={7976},
  pages={1089--1100},
  year={2023},
  publisher={Nature Publishing Group UK London}
}

@article{dauparas2022robust,
  title={Robust deep learning--based protein sequence design using ProteinMPNN},
  author={Dauparas, Justas and Anishchenko, Ivan and Bennett, Nathaniel and Bai, Hua and Ragotte, Robert J and Milles, Lukas F and Wicky, Basile IM and Courbet, Alexis and de Haas, Rob J and Bethel, Neville and others},
  journal={Science},
  volume={378},
  number={6615},
  pages={49--56},
  year={2022},
  publisher={American Association for the Advancement of Science}
}

@article{rives2021biological,
  title={Biological structure and function emerge from scaling unsupervised learning to 250 million protein sequences},
  author={Rives, Alexander and Meier, Joshua and Sercu, Tom and Goyal, Siddharth and Lin, Zeming and Liu, Jason and Guo, Demi and Ott, Myle and Zitnick, C Lawrence and Ma, Jerry and others},
  journal={Proceedings of the national academy of sciences},
  volume={118},
  number={15},
  pages={e2016239118},
  year={2021},
  publisher={National Academy of Sciences}
}

@article{lin2023evolutionary,
  title={Evolutionary-scale prediction of atomic-level protein structure with a language model},
  author={Lin, Zeming and Akin, Halil and Rao, Roshan and Hie, Brian and Zhu, Zhongkai and Lu, Wenting and Smetanin, Nikita and Verkuil, Robert and Kabeli, Ori and Shmueli, Yaniv and others},
  journal={Science},
  volume={379},
  number={6637},
  pages={1123--1130},
  year={2023},
  publisher={American Association for the Advancement of Science}
}

@article{merchant2023scaling,
  title={Scaling deep learning for materials discovery},
  author={Merchant, Amil and Batzner, Simon and Schoenholz, Samuel S and Aykol, Muratahan and Cheon, Gowoon and Cubuk, Ekin Dogus},
  journal={Nature},
  volume={624},
  number={7990},
  pages={80--85},
  year={2023},
  publisher={Nature Publishing Group UK London}
}

@article{stokes2020deep,
  title={A deep learning approach to antibiotic discovery},
  author={Stokes, Jonathan M and Yang, Kevin and Swanson, Kyle and Jin, Wengong and Cubillos-Ruiz, Andres and Donghia, Nina M and MacNair, Craig R and French, Shawn and Carfrae, Lindsey A and Bloom-Ackermann, Zohar and others},
  journal={Cell},
  volume={180},
  number={4},
  pages={688--702},
  year={2020},
  publisher={Elsevier}
}

@article{granda2018controlling,
  title={Controlling an organic synthesis robot with machine learning to search for new reactivity},
  author={Granda, Jaros{\l}aw M and Donina, Liva and Dragone, Vincenza and Long, De-Liang and Cronin, Leroy},
  journal={Nature},
  volume={559},
  number={7714},
  pages={377--381},
  year={2018},
  publisher={Nature Publishing Group UK London}
}

@article{haese2019olympus,
  title={Olympus: a benchmarking framework for noisy optimization and experiment planning},
  author={H{\"a}se, Florian and Aldeghi, Matteo and Hickman, Riley J and Roch, Lo{\"\i}c M and Christensen, Melodie and Liles, Elena and Hein, Jason E and Aspuru-Guzik, Al{\'a}n},
  journal={Machine Learning: Science and Technology},
  volume={2},
  number={3},
  pages={035021},
  year={2021},
  publisher={IOP Publishing}
}

@article{finn2017model,
  title={Model-Agnostic Meta-Learning for Fast Adaptation of Deep Networks},
  author={Finn, Chelsea and Abbeel, Pieter and Levine, Sergey},
  journal={arXiv preprint arXiv:1703.03400},
  year={2017}
}

@article{yosinski2014transferable,
  title={How transferable are features in deep neural networks?},
  author={Yosinski, Jason and Clune, Jeff and Bengio, Yoshua and Lipson, Hod},
  journal={Advances in neural information processing systems},
  volume={27},
  year={2014}
}

@article{cully2015robots,
  title={Robots that can adapt like animals},
  author={Cully, Antoine and Clune, Jeff and Tarapore, Danesh and Mouret, Jean-Baptiste},
  journal={Nature},
  volume={521},
  number={7553},
  pages={503--507},
  year={2015},
  publisher={Nature Publishing Group UK London}
}

@article{stanley2017openendedness,
  title={Evolution and the knightian blindspot of machine learning},
  author={Lehman, Joel and Meyerson, Elliot and El-Gaaly, Tarek and Stanley, Kenneth O and Ziyaee, Tarin},
  journal={arXiv preprint arXiv:2501.13075},
  year={2025}
}

@article{yao2023tree,
  title={Tree of thoughts: Deliberate problem solving with large language models},
  author={Yao, Shunyu and Yu, Dian and Zhao, Jeffrey and Shafran, Izhak and Griffiths, Tom and Cao, Yuan and Narasimhan, Karthik},
  journal={Advances in neural information processing systems},
  volume={36},
  pages={11809--11822},
  year={2023}
}

@article{madaan2023selfrefine,
  title={Self-refine: Iterative refinement with self-feedback},
  author={Madaan, Aman and Tandon, Niket and Gupta, Prakhar and Hallinan, Skyler and Gao, Luyu and Wiegreffe, Sarah and Alon, Uri and Dziri, Nouha and Prabhumoye, Shrimai and Yang, Yiming and others},
  journal={Advances in neural information processing systems},
  volume={36},
  pages={46534--46594},
  year={2023}
}

@article{schick2023toolformer,
  title={Toolformer: Language models can teach themselves to use tools},
  author={Schick, Timo and Dwivedi-Yu, Jane and Dess{\`\i}, Roberto and Raileanu, Roberta and Lomeli, Maria and Hambro, Eric and Zettlemoyer, Luke and Cancedda, Nicola and Scialom, Thomas},
  journal={Advances in neural information processing systems},
  volume={36},
  pages={68539--68551},
  year={2023}
}

@article{schmidhuber2003godel,
  title={G{\"o}del machines: self-referential universal problem solvers making provably optimal self-improvements},
  author={Schmidhuber, J{\"u}rgen},
  journal={arXiv preprint cs/0309048},
  year={2003}
}

@article{brest2006self,
  title={Self-adapting control parameters in differential evolution: A comparative study on numerical benchmark problems},
  author={Brest, Janez and Greiner, Sao and Boskovic, Borko and Mernik, Marjan and Zumer, Viljem},
  journal={IEEE transactions on evolutionary computation},
  volume={10},
  number={6},
  pages={646--657},
  year={2006},
  publisher={IEEE}
}

@article{krenn2022predicting,
  title={On scientific understanding with artificial intelligence},
  author={Krenn, Mario and Pollice, Robert and Guo, Si Yue and Aldeghi, Matteo and Cervera-Lierta, Alba and Friederich, Pascal and dos Passos Gomes, Gabriel and H{\"a}se, Florian and Jinich, Adrian and Nigam, AkshatKumar and others},
  journal={Nature Reviews Physics},
  volume={4},
  number={12},
  pages={761--769},
  year={2022},
  publisher={Nature Publishing Group UK London}
}

@article{li2017hyperband,
  title={Hyperband: A novel bandit-based approach to hyperparameter optimization},
  author={Li, Lisha and Jamieson, Kevin and DeSalvo, Giulia and Rostamizadeh, Afshin and Talwalkar, Ameet},
  journal={Journal of machine learning research},
  volume={18},
  number={185},
  pages={1--52},
  year={2018}
}

\end{document}